\documentclass[twoside]{article}

\usepackage[accepted]{aistats2022}
\usepackage[round]{natbib}

\usepackage{graphicx}
\usepackage{subfig}
\usepackage{gensymb}
\usepackage{dblfloatfix} 
\usepackage{wrapfig}
\usepackage{lscape}
\usepackage{booktabs}

\newcommand{\squishlisttwo}{
 \begin{list}{$\bullet$}
  { \setlength{\itemsep}{0pt}
    \setlength{\parsep}{0pt}
    \setlength{\topsep}{0pt}
    \setlength{\partopsep}{0pt}
    \setlength{\leftmargin}{1.5em}
    \setlength{\labelwidth}{1.5em}
    \setlength{\labelsep}{0.5em} } }
\newcommand{\squishend}{
  \end{list}  }
 
\newcommand{\eqmoveup}{\vspace{-0.0in}}  
\newcommand{\captionmoveup}{\eqmoveup\vspace{-0.14in}}
  
\usepackage{babel}
\usepackage{hyperref}

\UseRawInputEncoding

\begin{document}

%

%
\runningauthor{Semenova, Xu, Howes, Rashid, Bhatt, Mishra, Flaxman}

\twocolumn[

\aistatstitle{PriorVAE: Encoding spatial priors with VAEs for small-area estimation}

\aistatsauthor{Elizaveta Semenova$^{*\dagger}$ \And Yidan Xu$^*$ \And Adam Howes}
\aistatsaddress{University of Oxford \And University of Michigan \And Imperial College London}

\aistatsauthor{Theo  Rashid  \And Samir Bhatt  \And Swapnil Mishra  \And Seth Flaxman $^\dagger$}
\aistatsaddress{Imperial College London \And Imperial College London, \\University of Copenhagen \And Imperial College London, \\University of Copenhagen \And University of Oxford} 

]

\begin{abstract}
Gaussian processes (GPs), implemented through multivariate Gaussian distributions for a finite collection of data, are the most popular approach in small-area 
spatial statistical modelling. In this context they are used to encode correlation structures over space 
and can generalise well in interpolation tasks.  Despite their flexibility, off-the-shelf GPs present serious computational challenges which limit their scalability and practical usefulness in applied settings. Here, we propose a novel, deep generative modelling approach to tackle this challenge, termed PriorVAE: for a particular spatial setting, we approximate a class of GP priors through prior sampling and subsequent fitting of a variational autoencoder (VAE). Given a trained VAE, the resultant decoder allows spatial inference to become incredibly efficient due to the low dimensional, independently distributed latent Gaussian space representation of the VAE. Once trained, inference using the VAE decoder replaces the GP within a Bayesian sampling framework. This approach provides tractable and easy-to-implement  means of approximately encoding spatial priors and facilitates efficient statistical inference. We demonstrate the utility of our VAE two stage approach on Bayesian, small-area estimation tasks. \textbf{Keywords:} spatial modelling, Gaussian process prior, small-area estimation, Bayesian inference, variational autoencoder
\end{abstract}

\section{Introduction}

Spatially referenced data come in a variety of forms, including exact geographical coordinates such as a latitude and longitude or predefined geographical areal units such as a village, administrative unit or pixel of a raster image. The latter are known as areal unit data, and are found in fields such as epidemiology, environmental and political science; a variety of relevant methods come under the banner of small-area statistics \citep{rao2015small}. There are many motivations for modelling such data, from surveillance program evaluation to identifying environmental risk factors for disease. Small-area statistics are particularly relevant to informing policy decisions, which are often made at the areal unit level \citep{clements2006bayesian}.

Statistical modelling of spatial data is routinely performed using Gaussian process (GP) priors \citep{williams2006gaussian}. GPs have gained popularity in a variety of applied fields due to their flexibility, ease of implementation, and their inherent ability to characterise uncertainty. However, GPs also present a number of practical challenges. For example, basic inference and prediction using a GP requires matrix inversions and determinants - both of which scale cubicly with data size. This makes applications of GPs prohibitive for large datasets. Moreover, kernel design of a GP requires substantial domain knowledge in order to reflect characteristics of the process of interest \citep{stephenson2021measuring}. Hence, the choice of an inference method is of great importance when it comes to dealing with GPs. The theoretic asymptotic convergence properties and diversity of Markov chain Monte Carlo (MCMC) approaches make it the most reliable method for Bayesian inference. However, MCMC scales poorly, and struggles to deal with the high degree of auto-correlation inherent to spatial data, limiting its utility in large spatial settings \citep{rue2009approximate}.  A range of more tractable approximate methods have been developed, such as Laplace approximation, variational inference \citep{hoffman2013stochastic}, expectation propagation \citep{minka2013expectation}, or inducing variables in sparse GPs \citep{titsias2009variational}. However, few of these approximate methods have asymptomatic guarantees or yield consistently accurate posterior estimates \citep{yao2018yes} over a wide range of challenging datasets. User-friendly software packages, such as R-INLA \citep{martins2013bayesian} provide extremely convenient interfaces to a large set of predefined models, but do not provide enough flexibility for custom model development, and hence, have limitations for specific classes of applied research. There is an unmet need for approaches that can be easily implemented and customised in popular probabilistic programming languages such as Stan \citep{carpenter2017stan}, NumPyro \citep{phan2019composable}, PyMC3 \citep{salvatier2016probabilistic} or Turing.jl \citep{ge2018turing}, while still scaling favourably to large datasets. Here, we propose a novel computational technique which leverages the variational autoencoder model from deep learning and combines it with Bayesian inference~\citep{mishra2020pi, fortuin2020gp} with the goal of small-area estimation.

An autoencoder \citep{hinton2006reducing} is a neural network architecture used for the task of \emph{supervised} dimensionality reduction and representation learning. It is comprised of three components: an encoder network, a decoder network, and  a low dimensional layer containing a latent representation. The first component, the encoder $\mathcal{E}_{\gamma}(\cdot)$ with parameters $\gamma$,
maps an input $x \in \mathcal{R}^p$ into the latent variable $z \in \mathcal{R}^d$, where $d$ is typically lower than $p$. The decoder network $\mathcal{D}_{\psi}(\cdot)$ with parameters $\psi$ 
aims to reconstruct the original data $x$ from the latent representation $z$. Therefore, an autoencoder imposes a `bottleneck' layer in the network which enforces a compressed representation $z$ of the original input $x$. This type of dimensionality reduction technique is particularly useful when some structure exists in the data, such as spatial correlations. This structure can be learnt and consequently leveraged when forcing the input through the network's bottleneck. Parameters of 
the encoder and decoder networks
are learnt through the minimisation of a reconstruction loss function $p(x|\hat{x})$ expressed as the likelihood of the observed data $x$ given the reconstructed data $\hat{x}$.

A variational autoencoder (VAE) extends this formulation to a probabilistic generative model \citep{kingma2013auto}. Rather than learning a latent representation $z$, it learns a probabilistic distribution $p(z|x),$ from which a latent representation can be sampled. This distribution is chosen in such a way, that the latent variables $z$ are independent and normally distributed. To achieve this, the encoder outputs a pair of $d$-dimensional vectors $\mu_\gamma, \sigma^2_\gamma$ to characterise the mean and variance of $z,$ respectively. The latent variable then follows the Gaussian distribution $z \sim \mathcal{N}(\mu_\gamma, \sigma^2_\gamma I)$. The loss function is modified to include a KL-divergence term implied by a standard $\mathcal{N}(0,I)$ prior on $z$: $\mathcal{L}(x, \hat{x}) = p(x|\hat{x}) + \text{KL}(\mathcal{N}(\mu_\gamma, \sigma^2_\gamma I) || \mathcal{N}(0, I))$. The KL term can be interpreted as a regulariser ensuring parsimony in the latent space. New data can be generated  by sampling from the latent space with the trained decoder network. This feature of VAEs has led to a series of successful applications with the goal of generating new samples from the approximated distribution of real data \citep{liu2018constrained}.

In this paper we propose a novel use of VAEs, termed PriorVAE: we learn uncorrelated representations of spatial GP priors for a predefined spatial setting, and then use the trained decoder to perform Bayesian inference on new data. 

Our contributions can be summarised as follows.
 \squishlisttwo
      \item We introduce a two stage process for inference. First, we train a VAE to create an uncorrelated representation of complex spatial priors. Next,  we use the learnt latent distribution in the model instead of the GP prior to perform MCMC inference on new data, while keeping the trained decoder fixed.  
      \item We demonstrate the usage of VAE-priors on a range of simulated and real-life datasets to show their performance in small-area estimation tasks. 
 \squishend

The rest of this paper is organised as follows. In Section~\ref{section:methods}, we lay out the methodology of the two-stage approach. In Section~\ref{section:experiments}, we demonstrate the approach on synthetic and real datasets. In Section~\ref{section:discussion}, we conclude with a discussion and provide a broader outlook on the potential impact of this work. 

\section{Methods} \label{section:methods}

\subsection{Three types of spatial data}
There are three types of spatial data: areal (or lattice), geostatistical (or point-references) and point patterns. Models of all three of them rely on the notion of GPs and their evaluations in the form of the multivariate normal distribution. 
Areal data arise when a fixed domain is partitioned into a finite number of sub-regions at which outcomes are aggregated. Number of disease cases recorded per county, regardless of the exact locations where within the county the cases occurred or got recorded, constitutes an example of areal data. State-of-the-art models of areal data rely on `borrowing strength' from neighbours and use a hierarchical Bayesian formulation to do so. Widely used models of such data capture spatial structure via the adjacency matrix $A$. Each element $a_{ij}$ of it is either 1, if areas $B_i$ and $B_j$ are neighbours, and 0 if not.  
Point-referenced (geostatistical) data represents measurements of a spatially continuous phenomenon at a set of fixed locations. Number of cases recorded at a set of hospitals with known geographical position is an example of such data. Modelling  of the underlying continuous process relies on pair-wise distances between observed and unobserved locations. GPs in such models use continuous kernels.
Point pattern data consist of precise locations of events. An example of a point pattern is a collection of GPS coordinates of households of newly occurred malaria cases. One way of modelling point pattern data is to cover the entire study area with a fine computational grid and view each grid cell as a location. Modelling of the continuous underlying process is typically done using continuous GP kernels.
All of the three types of data can be modelled using the proposed novel approach. 

\subsection{Latent Gaussian models}

Suppose we are given outcome data $y_1, \ldots, y_n$, corresponding to a set of observed disjoint areas $\{B_i\}, i=1,\dots,n,$ covering the domain of interest $G = \cup_{i=1}^n B_i.$ These areas may be the cells of a computational grid, pixels of a raster image, or they may correspond to existing administrative units. Outcome data $\{y_i\}_{i=1}^n$ can represent either counts aggregated over an area, such as number of disease cases, continuous bounded data, such as disease prevalence (i.e. a number between zero and one), or continuous unbounded data, such as land surface temperature in degrees Celsius ($\degree C$). Hence, in applied fields, it is common to use generalised linear models to unify the modelling approach for all such types of outcome. Accounting for mixed effect model structure, their Bayesian hierarchical formulation can be expressed as follows
\begin{align}
    \theta &\sim p(\theta), \label{eq:hyperprior}\\
    f | \theta &\sim \text{GP}(\mu, \Sigma(\theta)), \label{eq:f_gp}\\
    \eta &= X \beta + f , \label{eq:lin+pred} \\
    y | \eta &\sim p(u^{-1}(\eta), \theta). \label{eq:obs_model}
\end{align}
Here, (\ref{eq:hyperprior}) describes the hyperparameters, $\theta$, of the model, $f$ in (\ref{eq:f_gp}) denotes the latent Gaussian field defined by mean $\mu$ and covariance $\Sigma(\theta)$, $X$ in (\ref{eq:lin+pred}) is the fixed effects design matrix (a set of covariates), $\beta$ are the fixed effects, and $\eta$ is the linear predictor combining fixed and random effects. Equation (\ref{eq:obs_model}) provides an observational model, where $u$ is a link function characterising the mean of the distribution (e.g. logit for binomial data, exponential for positive data). A common modelling assumption is that the observations $y_i$ are conditionally independent $p(y|f, \theta) = \prod_{i=1}^n p(y_i|\eta(f_i), \theta),$ and the spatial structure is captured latently by function $f$. It is common to choose a GP prior over $f$, and as a consequence, finite realisations $f_\text{GP}$ are jointly normally distributed with mean $\mu$ and covariance matrix $\Sigma$. Since $f_\text{GP}$ always enters the model via the linear predictor, without loss of generality (affine transformation) we can  consider $f_\text{GP}$ to have zero mean ($\mu=0$) and be distributed as
$f_\text{GP} \sim \mathcal{N}(0, \Sigma)$. Structure of the covariance matrix $\Sigma$ depends on the spatial setting of the problem and the model which we chose for the random effect. Once the model for $f_\text{GP}$ has been defined, the linear predictor can be computed as
\begin{equation} \label{eq:lin_pred}
    u(E[y | f_\text{GP}]) = X\beta + f_\text{GP},
\end{equation}
and then linked to the observed data via the likelihood. Unless the random effect is chosen to be trivial (i.e. i.i.d. set of variables resulting from $\Sigma=I$), the random effect $f_\text{GP}$ represents the computational challenge. Further we describe some options for spatial random effect priors in the context of small-area estimation, and propose a method to substitute its calculation at the inference stage with another variable, leading to increased inference efficiency.

\subsection{Models of areal data}


A widely adopted group of approaches relies on the neighbourhood/adjacency structure and defines $\Sigma$ based on the connectivity of the adjacency graph. Such methods leverage the tendency for adjacent areas to share similar characteristics. Conditional Auto-Regressive (CAR) and Intrinsic Conditional Auto-Regressive (ICAR) models were first proposed by \citet{besag1974spatial} and later extended to the Besag-York-Mollié (BYM) model \citep{besag1991bayesian}. The general form of the spatial prior for this group of models is
\begin{align*}
    \phi \sim \mathcal{N}(0, Q^{-1})
\end{align*}
where the precision matrix $Q$ defines which specific model is used.
Under the CAR model, random effect $f_\text{CAR}=\phi$ is defined by the prior $\phi \sim \mathcal{N}(0, \tau^{-1}R^{-})$. Here by $Q^{-}=\tau^{-1}R^{-}$ we denote generalised inverse \citep{james1978generalised} of the precision matrix $Q$. The precision matrix is calculated as $Q = \tau (D - \alpha A)$. Here $A$ is the adjacency matrix, and $D$ is a diagonal matrix, with elements $d_{ii}$ given by the total number of neighbours of area $B_i$. The parameter $\tau$ is the marginal precision, and the parameter $\alpha$ controls the amount of spatial dependence. For instance, $\alpha=0$ implies complete independence and i.i.d. random effects. Condition $|\alpha|<1$ ensures that the joint distribution of $\phi$ is proper \citep{gelfand2003proper}. It is not uncommon, however, to use $\alpha=1$ leading to the degenerate precision matrix (hence, the $Q^{-}$ notation) and the ICAR model. In practice, ICAR models are supplied with additional constraints, such as $\sum_{i=1}^n \phi \approx 0$ to enable inference. 
The BYM model includes both an i.i.d. random effect component $\phi_1 \sim \mathcal{N}(0, \tau_1^{-1}I)$ to account for non-spatial heterogeneity and an ICAR component $\phi_2 \sim \mathcal{N}(0, Q_2^- )$, $Q_2 = \tau_2R_2$, for spatial auto-correlation. Hence, the total random effect can be calculated as $f_\text{BYM} = \phi_1 + \phi_2.$ Some reparameterisations of BYM have been recently proposed in the literature \citep{riebler2016intuitive} to improve the interpretability of the inference. The advantage of the set of models presented above is that they take neighbourhood structure into account. However, neither the shapes (irregularity) or the sizes of areas are captured.

Another natural way to model areal data, especially gridded surfaces with small area sizes, is modelling covariance between areas as an approximately spatially continuous process, for example based on the pairwise distances between the centroids. Typical kernels for distance-based covariance matrices include squared exponential, exponential, periodic or Mat\'{e}rn.

\subsection{Variational Autoencoders (VAEs)} \label{sec:VAE}


An autoencoder is a neural network that is trained by unsupervised learning, with the aim of dimensionality reduction and feature learning. It consists of an encoder network, $\mathcal{E}_{\gamma}(\cdot)$  with parameters $\gamma$, a decoder network $\mathcal{D}_{\psi}(\cdot)$ with parameters $\psi$ and a bottleneck layer containing a latent vector $z$. The encoder maps input $x \in \mathcal{R}^p$ into the latent vector  $z \in \mathcal{R}^d$, and the decoder network maps $z$ back into the input space by creating a reconstruction of the original data $\hat{x}= \mathcal{D}_{\psi}(\mathcal{E}_{\gamma}(x))$. The network is trained by optimisation to learn reconstructions $\hat{x}$ that are close to the original input $x$. An autoencoder can borrow any neural network architecture, including multilayer perceptrons, convolutional or recurrent layers.
A VAE is a directed probabilistic graphical model whose posterior is approximated by a neural network, forming an autoencoder-like architecture. The goal of VAEs is to train a probabilistic model in the form of $ p(x, z) = p(x|z)p(z)$ where $p(z)=\mathcal{N}(0, I)$
is a prior distribution over latent variables $z$ and $p(x|z)$ is the likelihood function that
generates data $x$ given latent variables $z$. The output of the encoder network $\mathcal{E}_{\gamma}$ in VAEs is a pair of $d-$dimensional vectors $\mu_{\gamma}(x), \sigma^2_{\gamma}(x)$, which can be used to construct the variational posterior for latent variable $z$. The decoder (generator) network $\mathcal{D}_{\psi}$ tries to reconstruct the input by producing $\hat{x}$. In particular, the model can be summarised into the following
\begin{eqnarray}
\label{eq:VAE}
(\mu_\gamma(x), \sigma^2_\gamma(x)) &=& \mathcal{E}_{\gamma}(x)\\
z|x &\sim& \mathcal{N}(\mu_\gamma(x), \sigma^2_\gamma(x)I)\\
\hat{x}|z &\sim&  \mathcal{D}_{\psi}(z), \quad z \sim \mathcal{N}(0, I)
\end{eqnarray}

Neural network parameters $\gamma$ and $\psi$ are estimated as maximisers of the evidence lower bound (ELBO) $$\mathcal{L} = p(x|z,\gamma, \psi) - \text{KL}\left(\mathcal{N}(\mu_{\gamma}(x), \sigma^2_{\gamma}(x)I) \| \mathcal{N}(0,I)\right)$$
or its extensions \citep{li2016r}. The first term in ELBO is the reconstruction loss, measured by likelihood quantifying how well $\hat{x}$ and $x$ match. The second term is a Kullback-Leibler (KL) divergence which ensures that $z$ is as similar as possible to the prior distribution, a standard normal. It serves as a regulariser ensuring parsimony in the latent space and thus leads to uncorrelated variables in the latent space. New data can be generated  by sampling from the latent space with the trained decoder network. Once the numerically optimal network parameters $\hat{\psi}$ have been obtained, new realisations can be generated in two steps. As the first step, we draw from the standard normal distribution $z \sim \mathcal{N}(0,I)$, and as the second step, apply the deterministic transformation $\mathcal{D}_{\hat{\psi}}(z)$.

\subsection{The proposed method} \label{sec:proposed}

Most commonly, VAEs have been used in the literature to learn and generate observed data. We propose using spatial priors $x=f_\text{GP}$, evaluated at the required spatial configuration (a grid or neighbourhood structure) as training examples instead of observed data. The trained VAE then enables us to compute $\hat{x}=f_\text{VAE}$. Remarkably, unlike dealing with observed data, this approach does not have issues with the quality nor the quantity of the training examples. The amount of training data is unlimited since we can draw as many GP realisations for training as required. Similarly, there is no issue of data quality as we can create exact GP draws, free from noise, characteristic for any real-life observations.

To perform MCMC inference in a spatial model, we replace evaluation of the GP-prior $f_\text{GP}$ in the linear predictor (\ref{eq:lin_pred}) with the learnt prior $f_\text{VAE}$ at the inference stage:
\begin{equation*} \label{eq:lin_pred_vae}
    u(E[y | f_\text{VAE}]) = X\beta + f_\text{VAE}.
\end{equation*}
Drawing from the standard normal distribution $z \sim \mathcal{N}(0,I)$ with uncorrelated entries $z_i$ leads to a much higher efficiency as compared to the highly correlated multivariate normals $\mathcal{N}(0, \Sigma)$ with a dense covariance matrix $\Sigma$. As a consequence, computation time also decreases, especially for models where $d<p$.

\section{Results}
\label{section:experiments}

\paragraph{One-dimensional GP on regular and irregular grids.}


\begin{figure*}[!t]
\centering
  \includegraphics[width=0.8\textwidth]{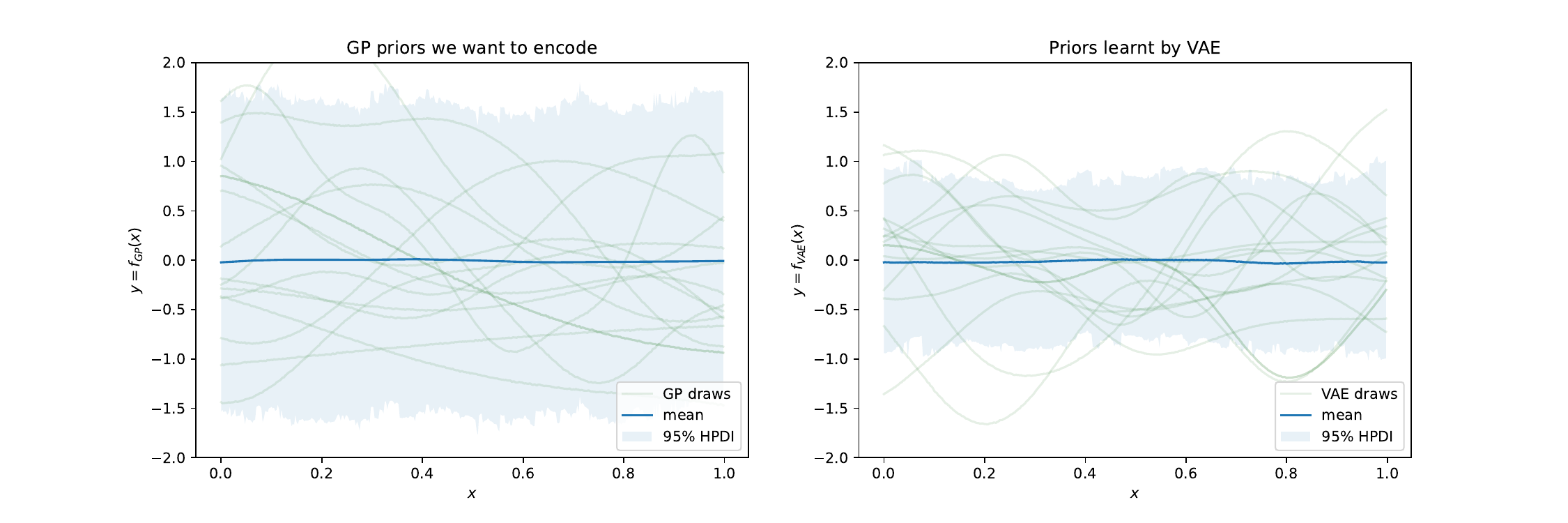}
  \caption{Learning one-dimensional GP-priors on a regular grid with VAE: left - prior samples from the original Gaussian process evaluations $f_\text{GP}$, right - prior draws from $f_\text{VAE}$ trained on $f_\text{GP}$ draws. Mean and highest posterior density interval (HPDI) are calculated from 1000 samples.}
  \label{fig:1}
\end{figure*}

\begin{figure*}[!t]
\centering
\hspace*{-2.5cm} 
  \includegraphics[width=21.5cm]{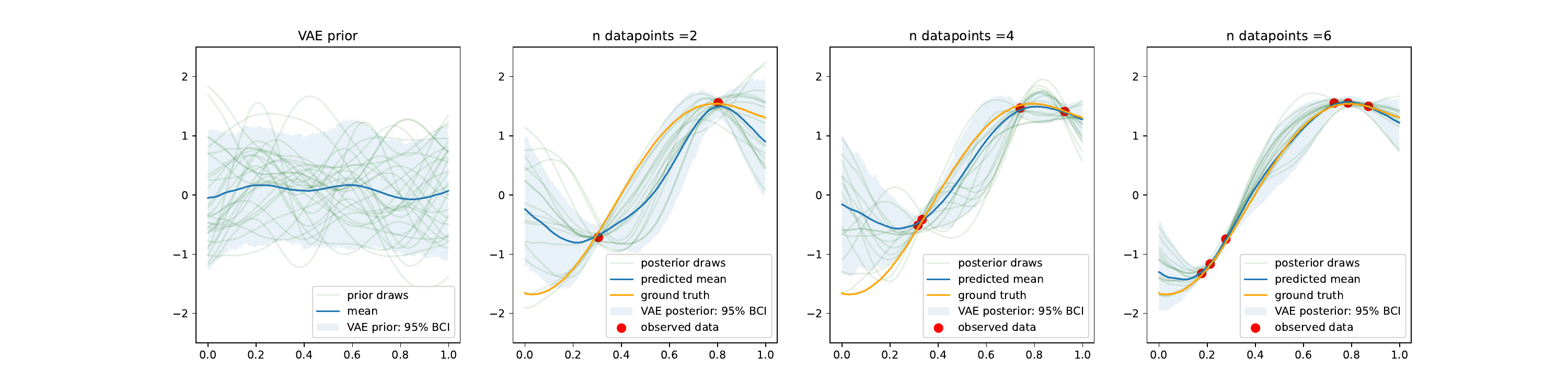}
  \caption{We perform MCMC inference on data generated from a noise-free GP and added i.i.d. noise by using the trained VAE priors $f_\text{VAE}$. The leftmost plot shows VAE prior. The plots on the right show posterior mean of our model in green with the $95\%$ credible intervals shown in blue. Quality of the estimation improves with the growing number of data points.}
  \label{fig:2}
\end{figure*}

\begin{figure}[htbp]
\centering
\hspace*{-2cm} 
  \includegraphics[height=6cm]{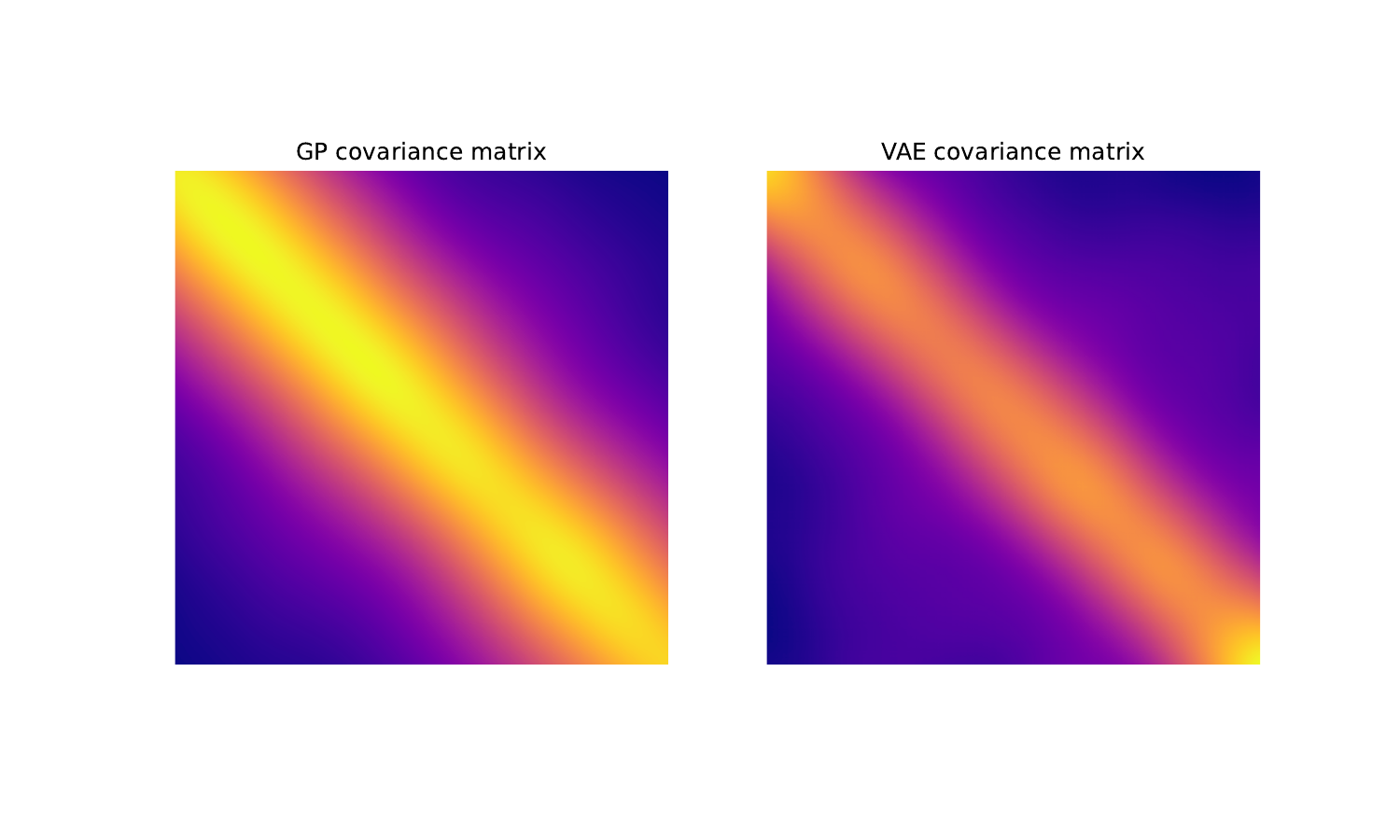}
  \caption{Covariance matrices computed empirically using 100 draws from the GP and VAE-GP priors.}
  \label{fig:cov_mats}
\end{figure}

In this first example we use VAE to perform inference on continuous data $\{y_i\}_{i=1}^n$ over a regular one-dimensional grid. The grid consists of $n=400$ points in the $(0,1)$ interval. Training prior samples are drawn as evaluations of a Gaussian process with zero mean and squared exponential kernel $k(h) = \sigma^2 e^{ -h^2/l^2}$. This model is useful for small-area estimation when the covariance matrix is Euclidean distance-based. To allow for hyperparameter learning, we impose hierarchical structure on the model by using hyperpriors on variance $\sigma^2 \sim \text{LogNormal}(0,0.1)$ and lengthscale $l \sim \text{InverseGamma}(4,1)$. This hierarchical structure allows the VAE to be trained on a range of values of hyperparameters. The average lengthscale, according to these priors, is around 0.3. Since this is much larger than the distance between two neighbouring points on the grid (0.0025), there is enough redundancy in the data to expect a lower dimensional embedding. Realisations $f_\text{GP}$ are presented on Figure \ref{fig:1}(left). We trained a VAE with two hidden layers of dimensions 35 and 30, respectively, and the bottleneck layer of dimension 10. As an activation function we used the rectified linear unit \citep{jarrett2009best, nair2010rectified, glorot2011deep} for all nodes in the hidden layers. The priors learnt by the VAE are presented in Figure~\ref{fig:1}(right). Figure \ref{fig:1} shows that the typical shape of the prior, as well as the mean have been learnt well. Amount of uncertainty (second moment) displayed by the VAE priors is lower. This is an expected phenomenon as vanilla VAEs are known to produce blurred and over-smoothed data due to compression; on the other hand, they perform well on denoising tasks (\cite{kovenko2020comprehensive}) i.e. reconstructing underlying truth given corrupt data. 
Empirical covariance matrices of the original and trained VAE priors show similar patterns (Supplement Figure \ref{fig:cov_mats}). 
To perform inference, we generate one GP realisation, use it as the ground truth, and simulate observed data by adding i.i.d. noise. We allow the number of observed data points to vary as 0.5\% (2 data points), 1\% (4 data points), and 1.5\% (6 data points) of the total number of the grid points and recover the true function. The model used for inference is $y \sim \mathcal{N}(f_\text{VAE}, s^2)$, where the amount of noise is given the half-Normal prior $s \sim \mathcal{N}^+(0.1)$.  Inference results are presented on Figure \ref{fig:2}. The higher the number of points, the closer is the estimated mean to the ground truth curve. Areas without any data available in their proximity, show higher uncertainty than areas informed by closely located data points. Effective sample size (ESS) is an important measure of the efficiency of MCMC sampling \citep{martino2017effective}. For example, we have run inference with 1000 warm-up and 1000 posterior MCMC samples for different number of data points. Average ESS for the posterior of the function evaluated at the observed points increased together with the number of points, while inference time remained constant. The original GP model displayed the reverse trend: average ESS remained constant, while computation time increased. 

To verify that our approach is also applicable to irregular grids, we modified the simulation study by placing the grid points irregularly. We use the same architecture to train this model, i.e. 35, 30 nodes in each hidden layer and 10 latent variables in the bottleneck layer. Original GP priors evaluated on an irregular grid, the learnt VAE priors and inference results are presented on Supplement Figures \ref{fig:irreg_draws} and \ref{fig:irreg_inference}. 

\begin{figure*}[!h]
\centering
\hspace*{-2.4cm} 
  \includegraphics[width=22.5cm]{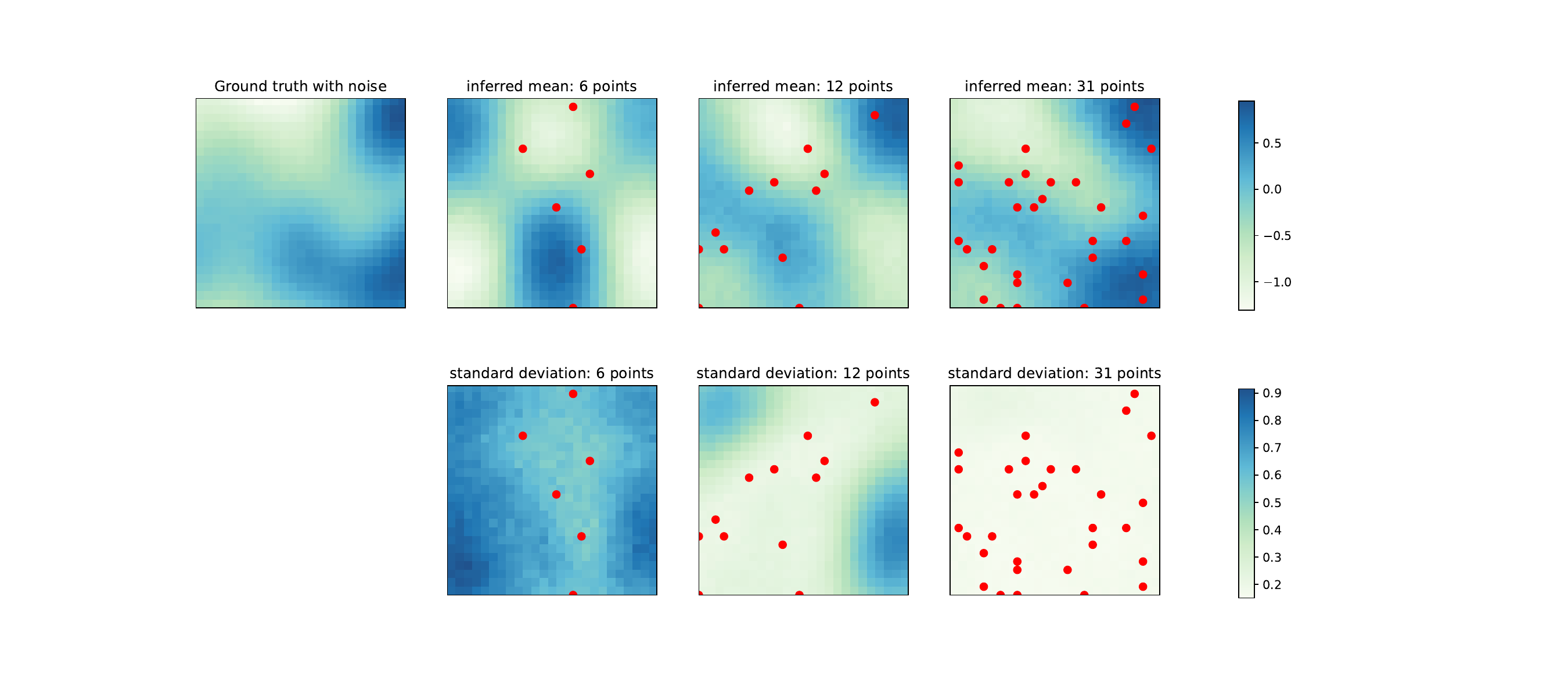}
  \caption{Inference results on a two-dimensional grid using the VAE-based priors. For comparison, inference has been performed using 1\%, 2\% and 5\% of the total number of points (6, 12, and 31 points, respectively). Uncertainty in the surface estimates decreases as the number of observed points increases.}
  \label{fig:2d_GP}
  \captionmoveup
\end{figure*}

\paragraph{Two-dimensional GP.} A similar set of experiments as described above were performed for two-dimensional GP priors. A regular grid was defined over a unit square with 25 segments along each of the two coordinates, resulting in $n=625$ grid cells. Inference results produced using a VAE trained on evaluations of two-dimensional GP priors are presented of Figure~\ref{fig:2d_GP}. As the number of observed points increases, quality of the mean prediction improves and uncertainty in the surface estimates decreases.

\paragraph{Synthetic CAR example.} We partition a rectangle by subdividing it into 10 rows and 15 columns. Unlike the two examples above, where we work with a continuous kernel, here we generate data using the CAR model $\phi \sim N(0,Q^{-1}), \quad Q = \tau(D-\alpha A),$ reliant on the adjacency matrix $A$. Parameter $\alpha$ is being assigned the $\text{Uniform}(0.4, 1)$ prior. This prior is chosen to study the question whether VAE can be trained well on CAR samples, displaying spatial correlation which corresponds to the values of $\alpha$ away from zero. 

In this example we demonstrate a technique to train a VAE by allowing it to preserve explicit estimation of the marginal precision $\tau$. For this VAE, the training is performed on the standardised samples $\bar{\phi}$ from the random effect, meaning that that the prior is drawn from the distribution
\begin{align}\label{eq:standardised}
\bar{\phi} \sim N(0, \bar{Q}^{-1}), \\
\bar{Q} = D- \alpha A \nonumber. 
\end{align}
Since $\phi = \frac{1}{\sqrt{\tau}} \bar{\phi} \sim N(0, Q^{-1})$, the trained VAE generator $\bar{f}_\text{VAE-CAR}$ also needs to be adjusted in the model at the inference stage by the magnitude of the marginal precision $\tau$: $f_\text{VAE-CAR} = \frac{1}{\sqrt{\tau}} \bar{f}_\text{VAE-CAR}.$

\begin{figure*}
\centering
\vspace*{-0.4cm} 
\hspace*{-1.4cm} 
  \includegraphics[height=8.0cm]{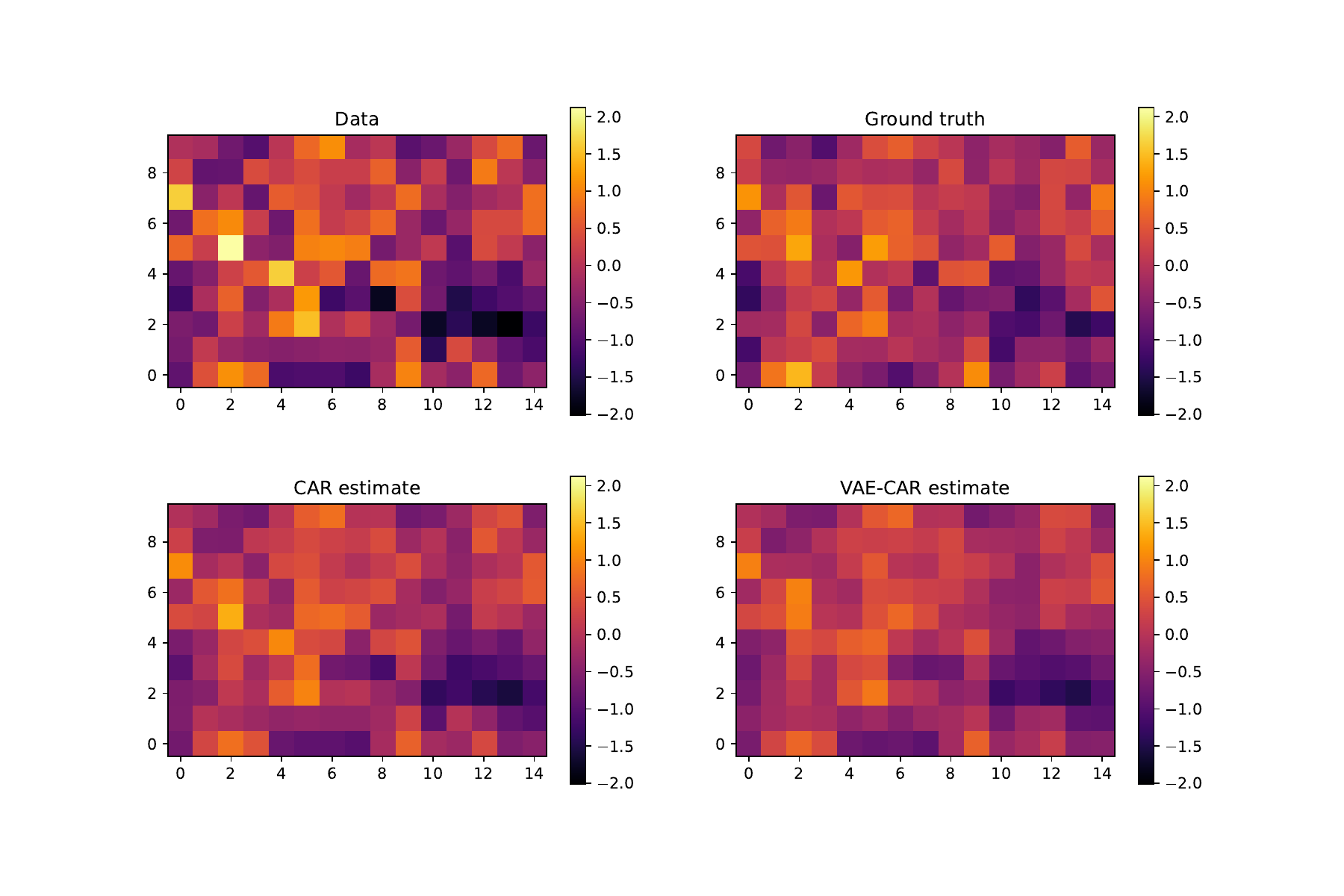}
  \caption{Inference results on a synthetic CAR example: data is generated by adding a noise term to the ground truth. We perform MCMC inference on the data by using both the original CAR and the VAE-CAR models.}
  \label{fig:synth_CAR}
  \captionmoveup
\end{figure*}

To test the performance of the trained VAE, ground truth data is generated by fixing the value of $\alpha$ at 0.7, and adding normal i.i.d. noise term at each area with variance of 0.5. This data is then used to fit the original CAR model, as well as the model using a VAE trained on CAR prior samples. Both models use the prior $\text{Uniform}(0.01, 1)$ for the variance of the noise term. For both models we run MCMC with 1000 warm-up iterations and 2000 iterations to sample from posterior. The average ESS of the spatial random effect in the VAE-CAR model is 4650 which took 9 seconds to compute; the achieved mean squared error (MSE) is 0.135. The average ESS in the original CAR model is 3128 which took 73 seconds; the achieved MSE is 0.118. Results of the experiment are presented on Figure \ref{fig:synth_CAR}.

\begin{figure}[h]
\hspace*{-1.1cm} 
\vspace*{-0.4cm} 
  \includegraphics[width=0.55\textwidth]{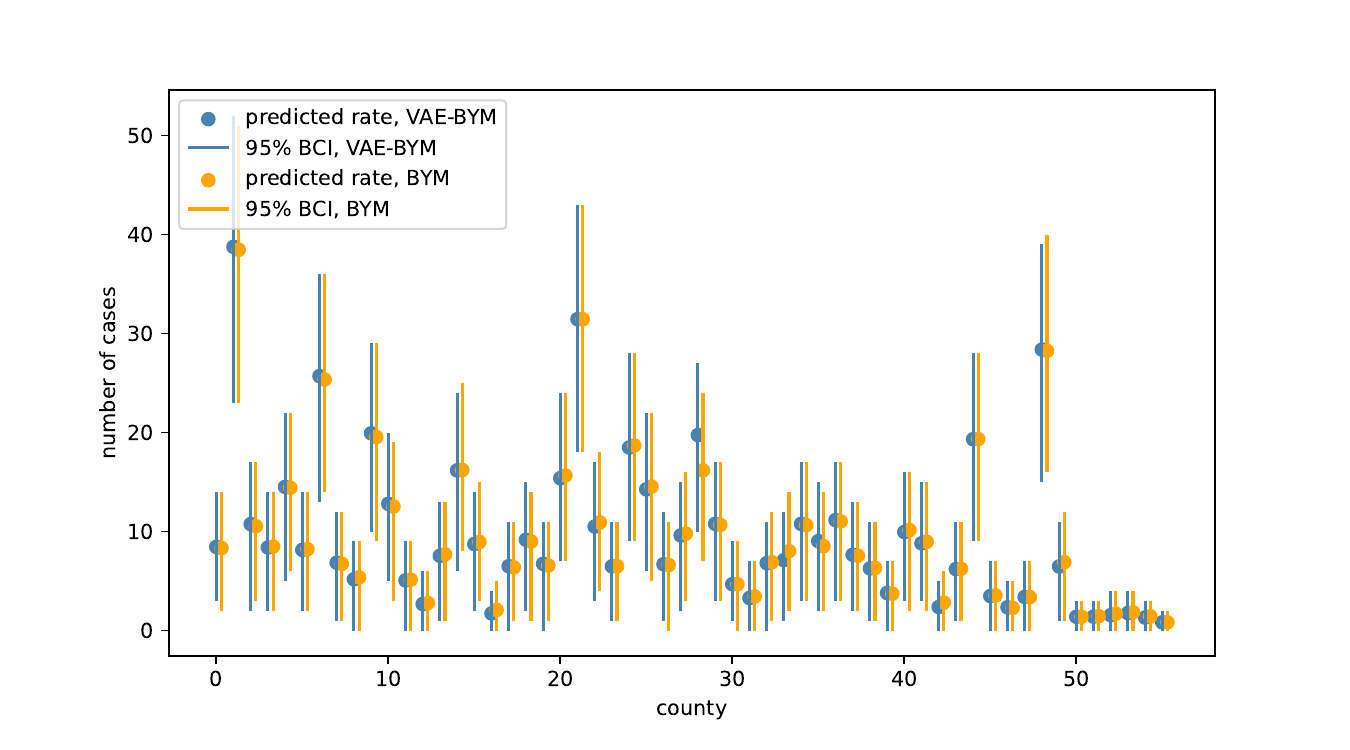}
  \caption{Scottish lip cancer dataset: posterior predictive distributions of the rates $\lambda_\text{BYM}$, $\lambda_\text{VAE-BYM}$ produced by models with $f_\text{BYM}$ and $f_\text{VAE-BYM}$ random effects, respectively. The distributions are represented by the point estimates (mean) and uncertainty intervals (95\% BCI) for each county. There is a very good agreement between the two models.}
  \label{fig:post_pred_BYM_VAEBYM}
\end{figure}

\begin{figure}[h]
\vspace{-0.4cm}
  \includegraphics[width=0.4\textwidth]{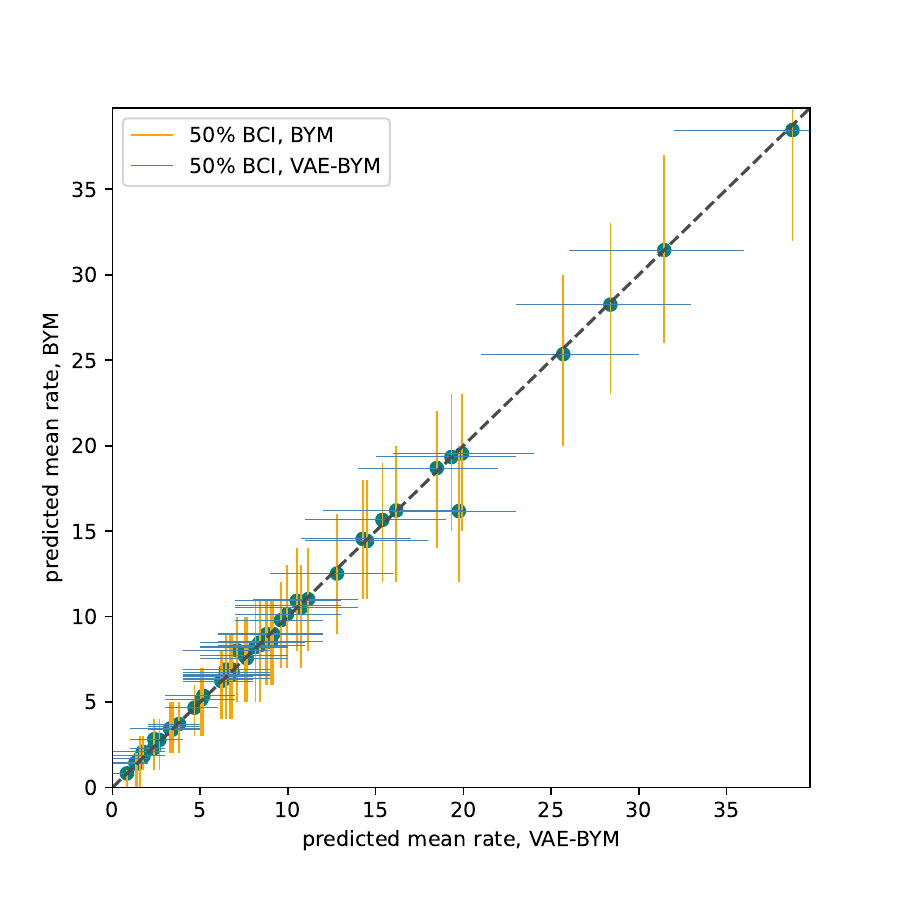}
  \caption{Scottish lip cancer dataset: point estimates (means) and uncertainty intervals (50\% BCIs) of the rates $\lambda_\text{BYM}$ and $\lambda_\text{VAE-BYM}$ produced by models with $f_\text{BYM}$ and $f_\text{VAE-BYM}$ random effects, respectively. There is a very good agreement between the two models.}
  \label{fig:scatter_BYM_VAEBYM}
\end{figure}

\begin{figure*}[t]
\centering
\hspace*{-3.1cm} 
  \includegraphics[width=22cm]{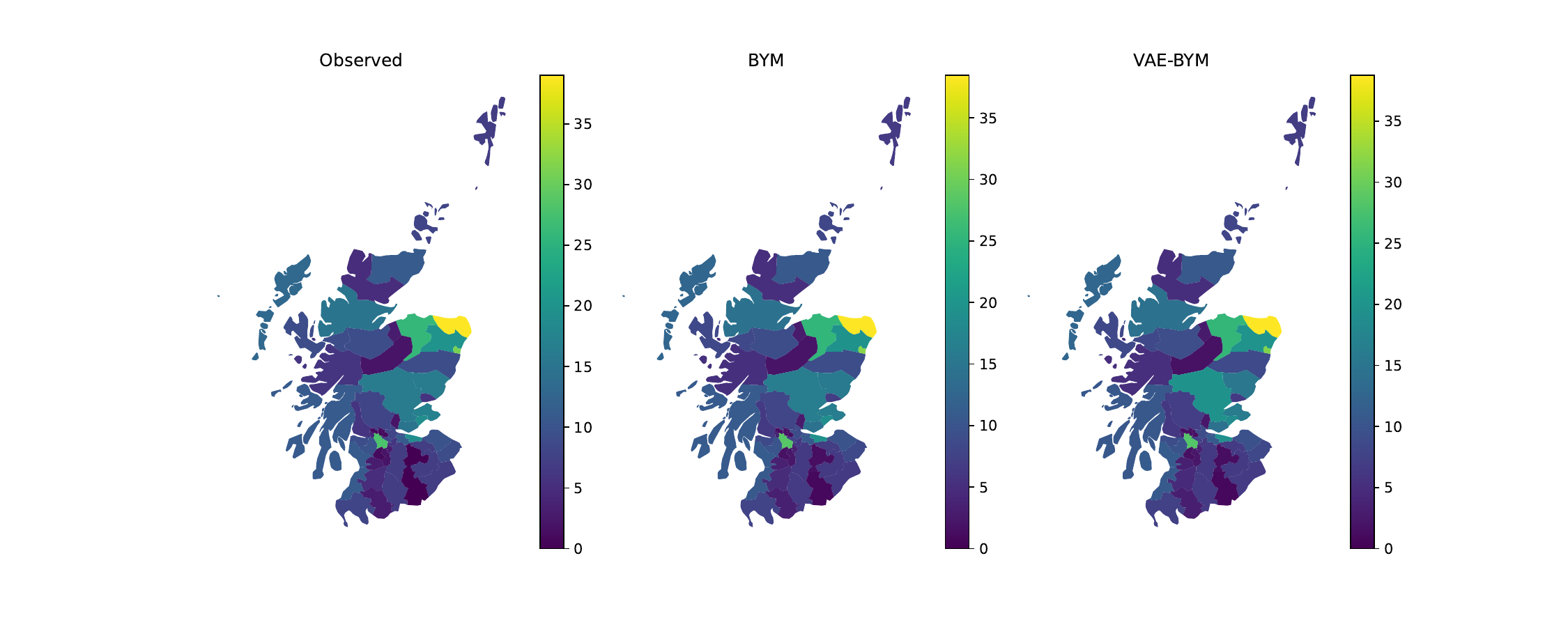}
  \caption{We perform MCMC inference on the observed number of lip cancer cases in each out of the 56 counties of Scotland. The leftmost plot displays the observed number of cases, the middle plot shows the mean of the posterior predictive distribution using the the BYM random effect $f_\text{BYM}$, the rightmost plot shows the mean of the posterior predictive distribution obtained from the model with the $f_\text{VAE-BYM}$ random effect. Models using BYM and VAE-BYM display the same pattern of the disease spatial distribution.}
  \label{fig:scotland_maps}
\end{figure*}

\paragraph{Scottish lip cancer dataset.} The Scottish lip cancer dataset, originally presented by \citet{kemp1985atlas}, has become a benchmark dataset for areal models. It has been used to demonstrate performance and implementations of CAR, ICAR, BYM and its variations \citep{duncan2017spatial, morris2019bayesian}. 
The dataset consists of the observed and expected numbers of cases ($y$ and $E$, respectively) at 56 counties in Scotland, as well as a covariate measuring the proportion of the population engaged in agriculture, fishing, or forestry (\textit{aff}). The covariate is related to exposure to sunlight, which is a risk factor for lip cancer. We model the count data $y$ as following a Poisson distribution with the log-Normal rate $\lambda$ distributed according to the BYM model: 
\begin{align}
y &\sim \text{Poisson}(\lambda), \nonumber\\ 
\log (\lambda) &= \log(E)+ b_0 + b_1 \text{\textit{aff}} + \phi_1 + \phi_2,\nonumber \\
\phi_1 &\sim  \mathcal{N}(0,\tau_1I), \nonumber \\
\phi_2 &\sim \mathcal{N}(0, Q_2^{-}),\quad Q_2 = \tau_2 (D - A). \nonumber
\end{align}
The VAE is trained on the spatial random effect BYM priors $f_\text{BYM} = \phi_1 + \phi_2$ to obtain the $f_\text{VAE-BYM}$ representation. We can use any parametrisation of BYM, as we only rely on its generative properties (and this is one of the advantages of our approach). We notice that a model with i.i.d. random effect $\phi_1$ already produces a relatively good fit (see Supplement, Figure \ref{fig:scottish_supplement}(a)). It is only the remaining discrepancies that the spatially-structured random effect $\phi_2$ needs to explain. We account for this observation in our priors for $\tau_1$ and $\tau_2$, as well as the dimension of the latent variable $z$: as there is only a moderate spatial signal, there is no redundancy in the data for spatial effect estimation. To be able to provide good quality VAE-BYM priors, we opt to not compress the spatial prior and choose the dimension of $z$ to be equal to the number of counties. We train a network with one hidden layer with 56 hidden nodes and use the exponential linear unit \citep{clevert2015fast}, or \textit{elu}, activation function. For optimisation, we use the variational R\'enyi bound that extends traditional variational inference to R\'enyi $\alpha$-divergences as proposed by \citet{li2016r}. By using $\alpha=0$ we opt for an importance weighted autoencoder \citep{burda2015importance}. We performed two assessments to evaluate whether the VAE-BYM produces similar inference results as the original BYM model. First, we used both models for inference on the entire dataset to compare results for mapping (the most typical task in epidemiology and policy informing work), and second, we performed 5-fold cross-validation. Posterior predictive distributions of the rates $\lambda_\text{BYM}$, $\lambda_\text{VAE-BYM}$  obtained by the models where $f_\text{BYM}$ and $f_\text{VAE-BYM}$ have been used to capture the spatial random effect, are very close to each other: Figure~\ref{fig:post_pred_BYM_VAEBYM} displays the two distributions, where each of them is represented by its point estimate (mean) and 95\% Bayesian credible interval (BCI). Uncertainty intervals produced by the two models are remarkably close to each other for most of the counties. Figure~\ref{fig:scatter_BYM_VAEBYM} demonstrates very good agreement in the point estimates. Figure~\ref{fig:scotland_maps} presents the obtained maps: models with $f_\text{BYM}$ and $f_\text{VAE-BYM}$ produce very close spatial patterns. The average ESS of the spatial effects in the BYM model is $\sim$150, and in the VAE-BYM model it is $\sim$1030. MCMC elapsed time shows the same trend: 402s and 12s for the BYM and VAE-BYM respectively. To perform cross-validation, we created a 5-fold split of the data. To measure performance, we have used MSE between the continuous predicted rate $\lambda$ and the observed count $y$. 
The mean MSE of the BYM model across five runs was 426, with standard deviation of 131, and the mean MSE of the VAE-BYM model was 414, with standard deviation of 171. Average ESS of the VAE-BYM random effect was $\sim$3850, and average ESS of the BYM random effect was $\sim$630. Inference times were 3s on average (0.2s standard deviation) for the VAE-BYM runs, and 33s on average (3s standard deviation) for the BYM runs. These experiments confirm the consistency of our observations: even when the dimension of the latent variable $z$ is the same as the number of the counties, there is a benefit to using VAE-BYM over BYM - it achieves comparable performance while displaying much higher ESS and shorter inference times.

\begin{figure}
\hspace*{-1cm} 
\vspace{-0.4cm}
  \includegraphics[width=0.55\textwidth, height=0.3\textwidth]{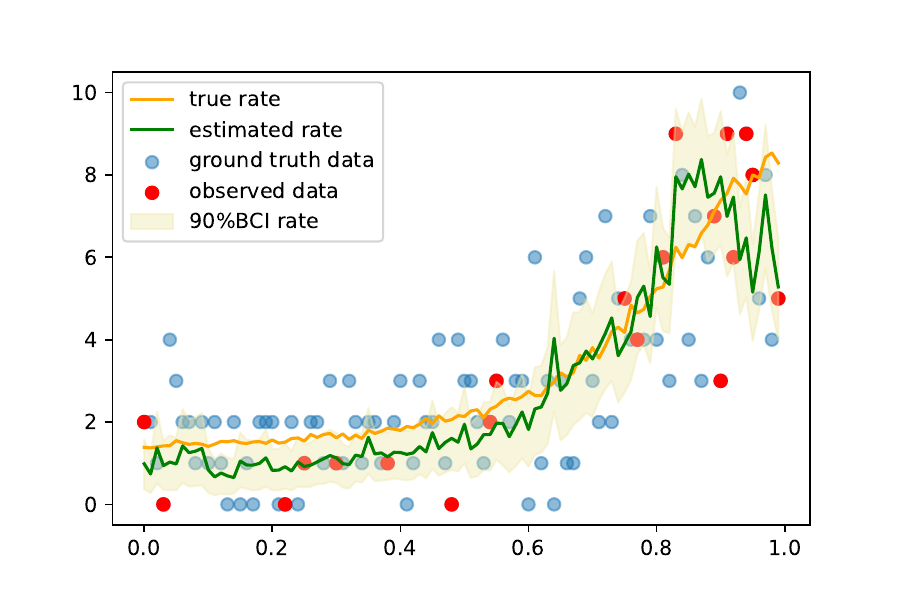}
  \caption{Inference results obtained via a VAE trained directly on the count data $y$ generated from the Poisson distribution $\text{Pois}(\lambda), \lambda=\exp(f_\text{GP})$. The estimated rate is in good agreement with the ground truth.}
  \label{fig:1d_Pois}
\end{figure}

\paragraph{An alternative approach to training a VAE for non-Gaussian likelihoods} Our previously described approach, where we calculate the linear predictor and then fit model to the data using a link function, follows the long standing tradition in statistics driven by the interest to measure associations between predictors and an outcome. In the previous two examples we have trained the VAE directly on the GP draws ($f_\text{GP}$) and used the Gaussian distribution to compute the reconstruction loss $p(x|\hat{x})$. When the outcome data is not Gaussian or not even continuous, there is an alternative way of training a VAE. Let us consider, for instance, count data which we would like to model using the Poisson distribution. Instead of using $f_\text{GP}$ draws as training data of the VAE, we can use directly the simulated counts $y$ arising from the Poisson distribution with rate $\lambda=\exp(f_\text{GP})$. The encoder now maps these counts into the latent variable $z$, as, before, and the decoder is enhanced with a probabilistic step: it maps $z$ into a positive rate $\hat{\lambda}$, and calculates the reconstruction loss as $p(y|\hat{\lambda}): y\sim \text{Pois}(\hat{\lambda})$. Results of such a model are shown on Figure~\ref{fig:1d_Pois}. We have generated training data $y$ for the VAE over a regular grid with $n=100$ points according to the distribution $y\sim \text{Pois}(\exp(f_\text{GP}))$. The hyperparameters of the GP were identical to our one-dimensional GP example presented above. VAE architecture included two hidden layers with \textit{elu} activation function. An exponential transform was applied to the final layer of the decoder to calculate the predicted rate $\hat{\lambda}$.

\begin{figure*}
\centering
\hspace*{-3.1cm} 
  \includegraphics[width=22cm]{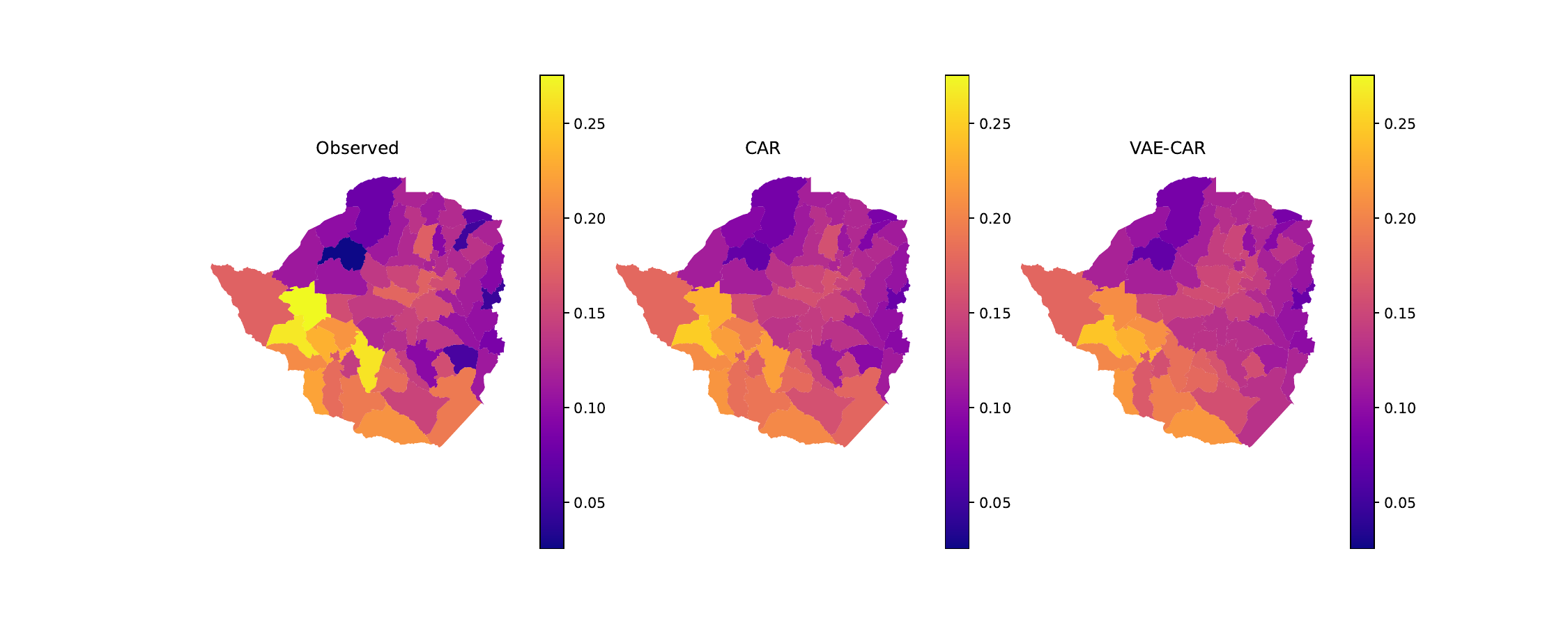}
  \caption{We perform MCMC inference on the observed number of HIV-positive individuals among observed individuals in Zimbabwe. The leftmost plot displays raw data, the middle plot shows the mean of the posterior predictive distribution using the the CAR spatial random effect $f_\text{CAR}$, the rightmost plot shows the mean of the posterior predictive distribution obtained from the model with the $f_\text{VAE-CAR}$ spatial random effect.}
  \label{fig:zimbabwe_maps}
\end{figure*}

\begin{figure*}
\centering
\hspace*{-1.5cm} 
\vspace*{1.5cm} 
  \includegraphics[width=20cm]{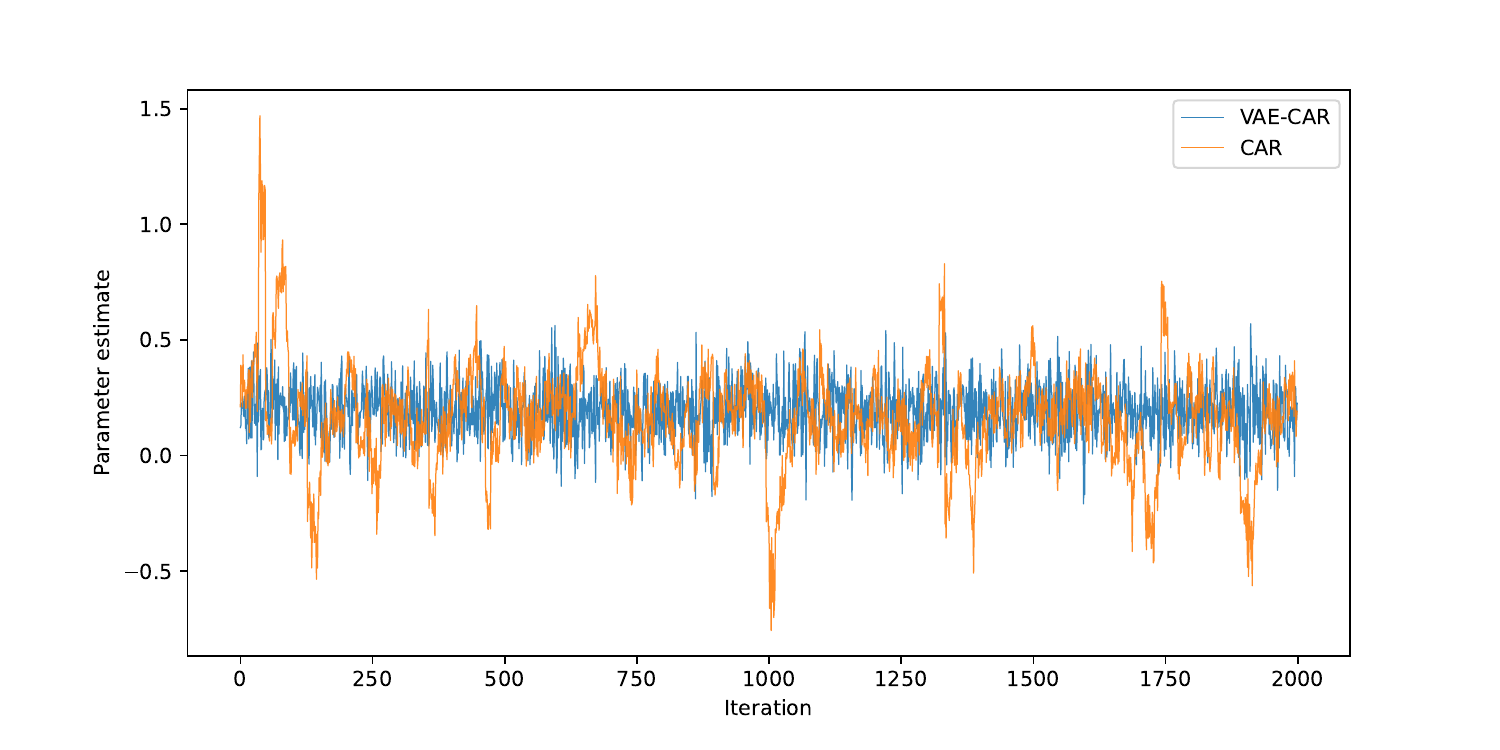}
  \caption{Traceplots of the spatial random effect for one area obtained during MCMC inference on HIV prevalence data from Zimbabwe. Posterior samples produced by the VAE-CAR model display very little auto-correlation, which allows to achieve high ESS and fast inference. Posterior samples produced by the original CAR model, on the contrary, show high auto-correlation, leading to low ESS and longer computation time.}
  \label{fig:zimb_traceplots}
  \captionmoveup
\end{figure*}

\paragraph{HIV prevalence in Zimbabwe.}
We consider household survey data from the 2015-2016 Population-based HIV Impact Assessment Survey in Zimbabwe \cite{sachathep2021population}.
The observed positive cases $y_i$ among all observed cases $n_i$ in each area $B_i$ are modelled according to the Binomial distribution: 
\begin{align}
y &\sim \text{Binomial}(n, \theta), \nonumber\\ 
\text{logit}^{-1} (\theta) &= b_0 + \phi,\nonumber \\
\phi &\sim  \mathcal{N}(0, Q^{-1}), \nonumber \\
Q &= \tau  (D - \alpha A). \nonumber
\end{align}
The parameter $\theta_i$ is the "probability of success" of the binomial distribution, and serves as the estimate of HIV prevalence. The linear predictor $\text{logit}^{-1} (\theta_i) = b_0 + \phi_i, \quad i=1,...,N$ consists of the intercept $b_0$, common for each area, and area-specific spatial random effect $\phi_i$. The VAE is trained on the spatial random effect CAR priors to obtain the VAE-CAR representation. We train VAE on the standardised draws of the spatial random effect as presented in formulas \ref{eq:standardised}.
The total number of observed areas is 63, and we trained a VAE with a dimension of 50 in the latent space to achieve reduction in the dimensionality of the latent variable. Figure~\ref{fig:zimbabwe_maps} presents the obtained maps: models with $f_\text{CAR}$ and $f_\text{VAE-CAR}$ produce very close spatial patterns. We used 2000 MCMC iterations to perform inference using both the original CAR, as well as the VAE-CAR models. The average ESS of the spatial effects in the CAR model is $\sim$120 with a computation time of 13 seconds. In the VAE-CAR model average ESS is higher, at $\sim$2600, and took only 4 seconds to compute. 
ESS achieved by the VAE-CAR model is about 20 times higher than the one achieved by the CAR model, and also higher than the actual number of posterior samples, which indicates extremely efficient sampling. Traceplots obtained by both models are presented on Figure \ref{fig:zimb_traceplots} and demonstrate (together with Supplement Figure \ref{fig:zimb_autocorr_vae}) that posterior samples produced by the VAE-CAR model display very little auto-correlation, which allows to achieve high ESS and fast inference. Posterior samples produced by the original CAR model, on the contrary, show high auto-correlation (\ref{fig:zimb_autocorr_car}), leading to low ESS and longer computation time.

\paragraph{COVID-19 incidence in the UK.}
We consider the projected number of COVID-19 infections in the UK at Lower Tier Local Authority (LTLA) level (\cite{mishra2020covid}). These estimates are available from the public website
\footnote{\url{https://imperialcollegelondon.github.io/covid19local/\#downloads}} and are based on the \cite{epidemia} package. The model behind the package (\cite{flaxman2020estimating}) relies on a self-renewal equation and does not take spatial correlations into account. Here we demonstrate how spatial modelling using the proposed approach can be performed on the incidence data. Number of infections $y_i$ in each LTLA during the period between the 21st of January and the 5th of February 2022 is modelled via the Poisson distribution according to the model
\begin{align}
y &\sim \text{Poisson}(\lambda), \nonumber\\ 
\log (\lambda) &= b_0 + \phi,\nonumber \\
\phi &= \frac{1}{\sqrt{\tau}}\bar{\phi}, \nonumber \\
\bar{\phi} &\sim  \mathcal{N}(0, \bar{Q}^{-1}), \nonumber \\
\bar{Q} &=   D - \alpha A. \nonumber
\end{align}

\begin{figure*}
\centering
\hspace*{-3.1cm} 
  \includegraphics[width=22cm]{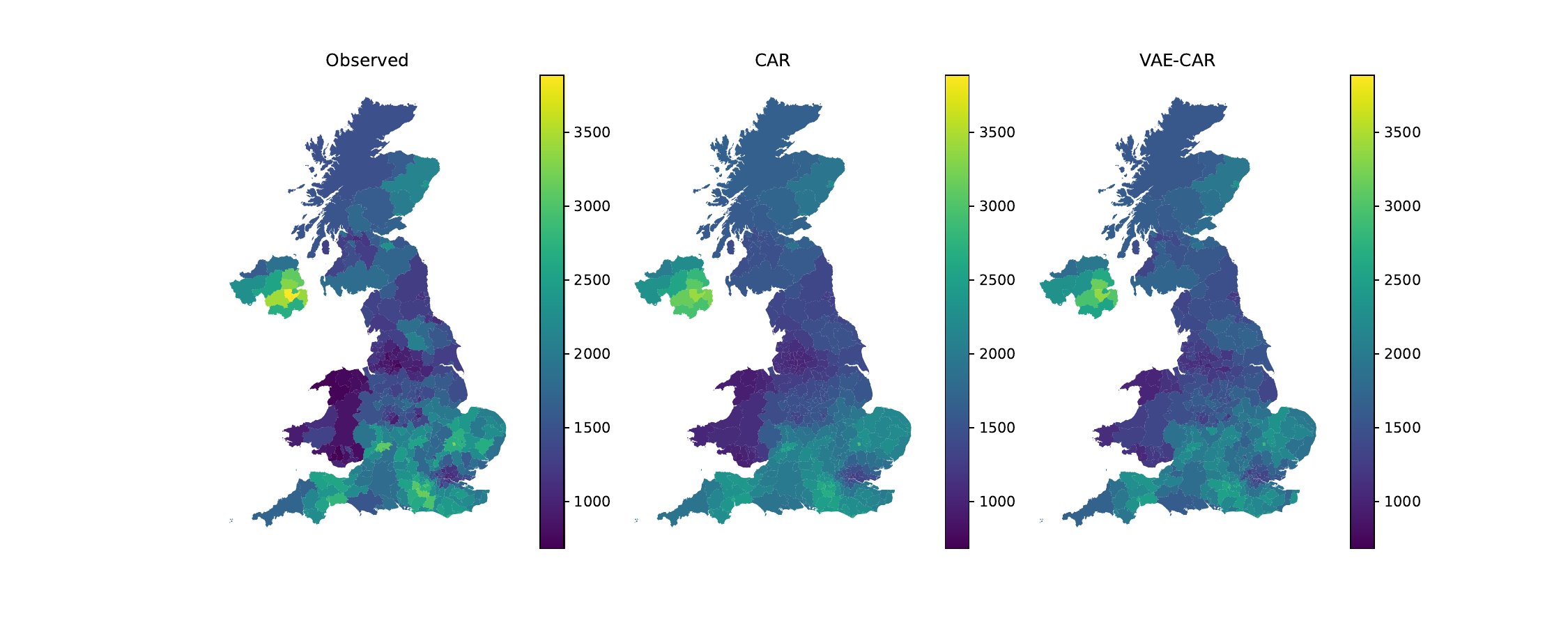}
  \caption{We perform MCMC inference on the projected COVID-19 incidence in the UK at the LTLA level. The leftmost plot displays raw data, the middle plot shows the mean of the posterior predictive distribution using the the CAR spatial random effect $f_\text{CAR}$, the rightmost plot shows the mean of the posterior predictive distribution obtained from the model with the $f_\text{VAE-CAR}$ spatial random effect.}
  \label{fig:ltla_maps}
\end{figure*}

As above, we have chosen the representation of the spatial CAR random effect in its standardised form $\bar{\phi}$ in order to allow for explicit inference of the marginal precision $\tau$ when using VAE-based inference. To simplify modelling, we have removed all singletons (islands without any neighbouring areas within them) from the shapefile. We trained VAE on the standardised draws of the spatial random effect. The total number of modelled LTLAs is 372, and we chose the dimension of the latent space to be 300. Using this example, we demonstrate a technique, where several VAEs using different priors for the spatial hyperparameter $\alpha$ can be pre-trained, and then used for model selection - a step enabled by fast inference times when using VAE-based models. We have trained five VAEs using hyper-priors $\alpha\sim U(0, 0.2)$, $U(0.2, 0.4)$, $U(0.4, 0.6)$, $U(0.6, 0.8)$, $U(0.8, 0.99)$, correspondingly. Each of the resulting VAE-priors were used to fit a model. Model selection was performed based on the widely applicable information criterion (WAIC; \cite{watanabe2010asymptotic}) using the \textit{arviz} package (\cite{arviz_2019}). The best model was the one trained with $\alpha\sim U(0.8, 0.99)$. To obtain smooth maps, we used 2000 MCMC iterations and performed inference using both the original CAR, as well as the VAE-CAR models with priors $\alpha \sim \text{Uniform}(0.8, 0.99)$ and $\tau \sim \text{Gamma}(6,2)$. Resulting maps are presented on Figure~\ref{fig:ltla_maps}: models with $f_\text{CAR}$ and $f_\text{VAE-CAR}$ produce similar spatial patterns. Characteristics of the two model fits are presented in Table~\ref{table:ltla_fit}: there is no significant difference between the absolute errors (p-value of the paired t-test is 0.05), while VAE-CAR has achieved much higher ESS at much shorter computation times. 


\begin{table}
\centering
\resizebox{\columnwidth}{!}{%
\begin{tabular}{ |c|c|c|c|c| } 
 \midrule
 Model  & \shortstack{Absolute error$^*$} &\shortstack{  ESS \\of the  spatial \\ random effect $^*$ }   & \shortstack{ MCMC \\elapsed\\ time (s)} \\ 
 \midrule
 CAR  & 1.53 (0.06) & 317 (6) & 277 \\ 
 VAE-CAR  & 1.63 (0.06) & 3188 (24) & 8\\ 
 \midrule
\end{tabular}%
}
\caption{Results of MCMC inference with CAR and VAE-CAR models: after 2000 iterations, VAE-CAR models achieve much higher average ESS at much shorter computation time, while displaying similar goodness of fit to the original model.\\
$^*$ Mean and standard error}
\label{table:ltla_fit}
\end{table}


\section{Discussion and Future work} \label{section:discussion}

In this paper we have proposed a novel application of VAEs to learn spatial GP priors and enable small-area estimation. Such an approach leverages the power of deep learning to fuel inference for well-established statistical models. Uncorrelated latent parameters of the VAE make consequent Bayesian inference with MCMC highly efficient at successfully exploring the posterior distribution. 
An advantage of the proposed approach, as compared to traditional VAE applications, is that there is no limitations in neither quality nor quantity of the training data as any number of training samples can be generated by drawing from the noise-free spatial priors. In addition, there is no need to retain the whole training data set as every batch can be generated on the fly.  Our method is beneficial even if the latent space dimension is the same as the number of data points, unlike some other approximation methods, which rely on information redundancy in the data. As the HIV prevalence in Zimbabwe example shows, by reducing the auto-correlation between posterior MCMC samples, we can obtain drastic gains in both ESS and computation speed.

The limitations of the proposed approach are as follows. Firstly, MCMC inference is restricted to the spatial structure used for training. For instance, if a fixed spatial grid or a neighbourhood structure was used to train the VAE, prediction for off-grid points would not be possible. Secondly, using the MSE loss for VAE training, we do not expect the VAE to work well for values of hyperparameters outside of the typical ranges of hyperpriors which the VAE was trained on: if a VAE was trained on GP realisations with short lengthscales, it is unreasonable to expect good results for long lengthscales. VAE training losses which capture distributional properties of the training samples have the potential to resolve it and constitute future work. The over-smoothing property of vanilla VAEs can be accounted for at the inference stage via priors allowing for wider uncertainty ranges. There is also the upfront cost of training a VAE, including the choice of architecture. Finally, recovering interpretable hyperparameters of the spatial priors might be of scientific interest and is more challenging in the VAE approach rather than the traditional framework. In some cases this difficulty can be alleviated. For instance, in the HIV prevalence example we have demonstrated how marginal precision or variance can be isolated from the VAE training to retain direct inference of this parameter.  


Statistical models of areal data described above assign the same characteristics to each location within every area $B_i$. This assumption is unrealistic for many settings as heterogeneity might be present within each $B_i$, as long as the size of the area is non-negligible. Areal data can be viewed as an aggregation of point data and a series of approaches began to emerge recently to address this issue (\cite{johnson2019spatially, arambepola2022simulation}). Hence, it is reasonable to use a data augmentation step to sample from the posterior distribution of the exact locations, and then to aggregate results. Future directions of research include an approach where presented above models, reliant on the adjacency structure, are substituted with continuous kernel approaches and a VAE trained using them.

The application of VAEs to small-area estimation has potential for far-reaching societal impact. If a set of different GP priors, such as CAR, ICAR, BYM and others, are used to pretrain several VAEs over the same spatial configuration, the resulting decoders can then be applied via the proposed inference scheme to rapidly solve real-life problems. Once the VAE training has been performed, users will only need access to the decoder parameters, and otherwise perform inference as usual – the expensive step of training a VAE would not be required at practitioner's end. In case of an epidemic emergency, for instance, this would enable faster turnaround times in informing policy by estimating crucial quantities such as disease incidence or prevalence.

\subsubsection*{Author contributions}
ES: formal analysis, methodology, software, visualization, writing of original draft, review and editing; YX: formal analysis, software, visualization, review and editing; AH: data curation, methodology, review and editing; TR: data curation, review and editing; SB: conceptualization, methodology, supervision, review and editing; SM: conceptualization, methodology, supervision, review and editing; SF: conceptualization, funding acquisition, methodology, supervision, review and editing.


\subsubsection*{Acknowledgements}
We thank Jeffrey Eaton for his useful comments on the manuscript. 


\subsubsection*{Funding Statement}

ES and SF acknowledge the EPSRC (EP/V002910/1). AH acknowledges EPSRC Centre for Doctoral Training in Modern Statistics and Statistical Machine Learning (EP/S023151/1). TR acknowledges Imperial College President's PhD Scholarship. SB acknowledges the MRC (MR/R015600/1), The Danish National Research Foundation via a chair position, and The NIHR Health Protection Research Unit in Modelling Methodology. SM and SB acknowledge funding from the Novo Nordisk Young Investigator Award (NNF20OC0059309).


\subsubsection*{Code availability}

Code is available at \url{https://github.com/elizavetasemenova/priorVAE}

\onecolumn
\aistatstitle{``PriorVAE: Encoding spatial priors with VAEs for small-area estimation" \\
Supplementary Materials}

\vspace{-0.5cm}

\textbf{Authors:}

Elizaveta Semenova (University of Oxford),\\
Yidan Xu (University of Michigan), \\
Adam Howes (Imperial College London), \\
Theo Rashid (Imperial College London), \\
Samir Bhatt (Imperial College London, University of Copenhagen), \\
Swapnil Mishra (Imperial College London, University of Copenhagen), \\
Seth Flaxman (University of Oxford)

\thispagestyle{empty}

\section*{One-dimensional synthetic example: regular grid}





\section*{One-dimensional synthetic example: irregular grid}
\begin{figure}[htbp]
   \centering
   \vspace*{-0.1cm} 
  \subfloat[] {
  \includegraphics[width=0.35\textwidth]{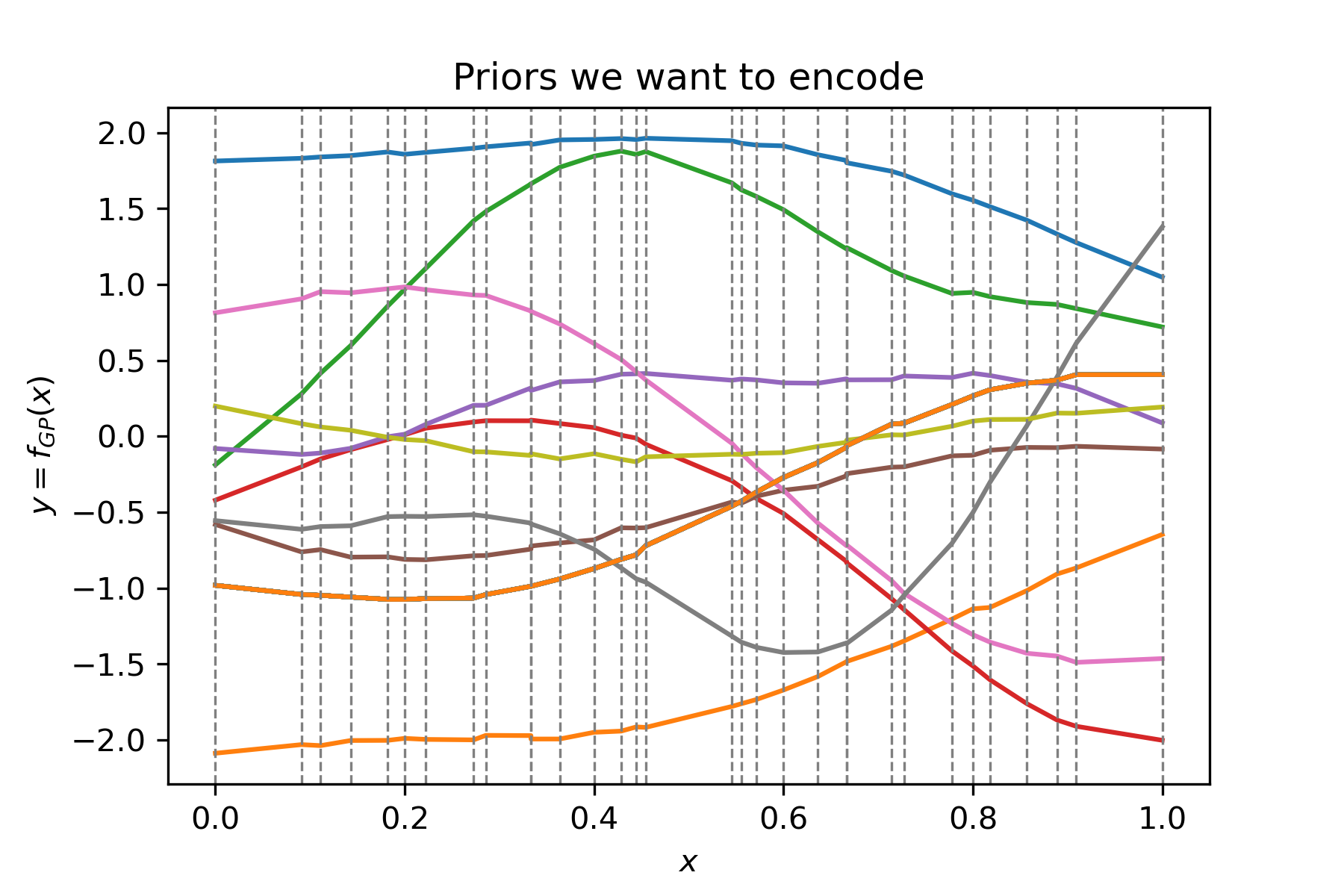} 
} 
\subfloat[] {
  \includegraphics[width=0.35\textwidth]{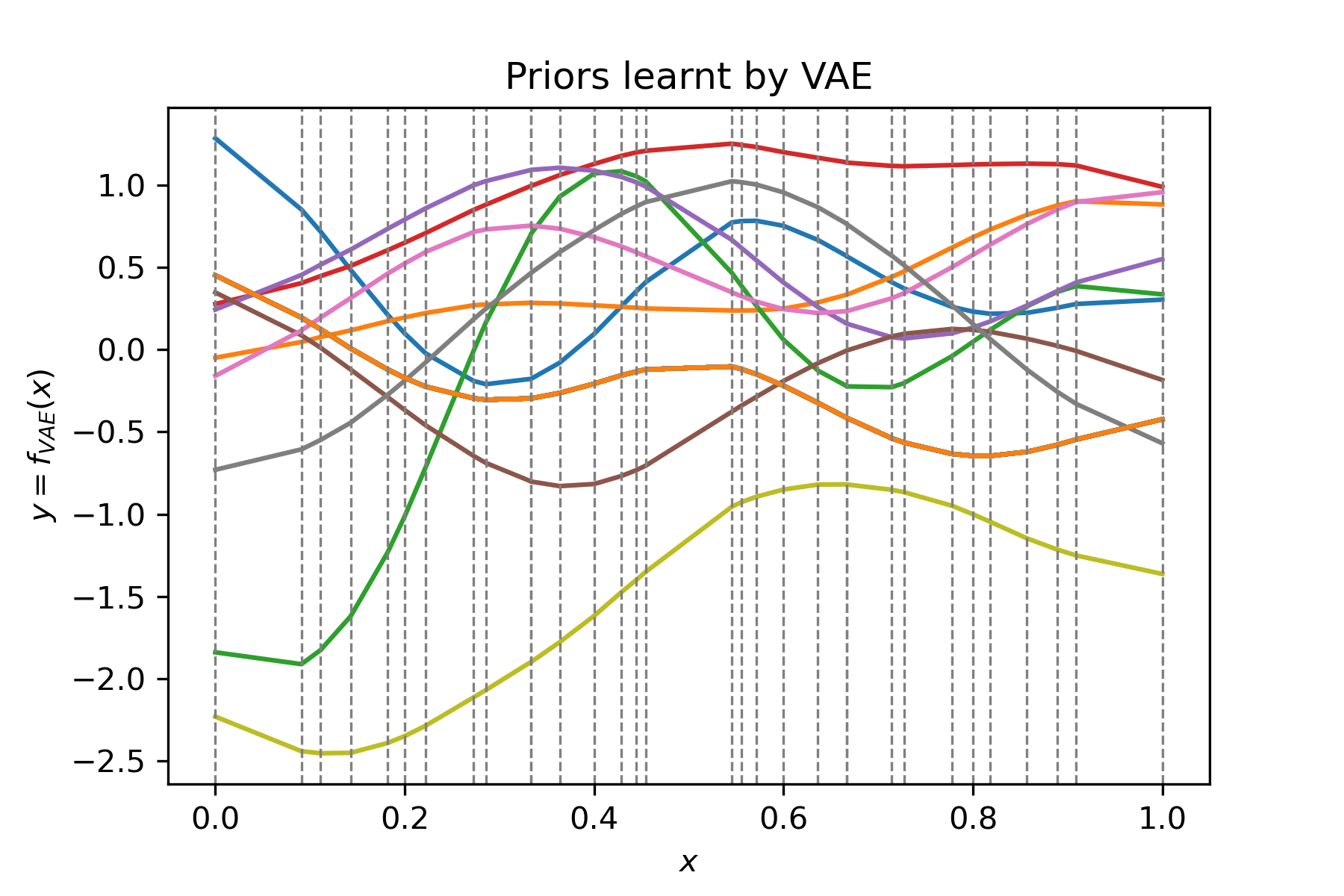}
} 
 \caption{Learning one-dimensional GP-priors on an irregular grid with VAE: (a) prior samples from the original Gaussian process evaluations $f_\text{GP}$, (b) prior draws from $f_\text{VAE}$ trained on $f_\text{GP}$ draws}
 \label{fig:irreg_draws}
\end{figure}

\begin{figure}[htbp]
\centering
\hspace*{-2cm} 
\vspace*{-0.2cm} 
  \includegraphics[width=21cm, height=4cm]{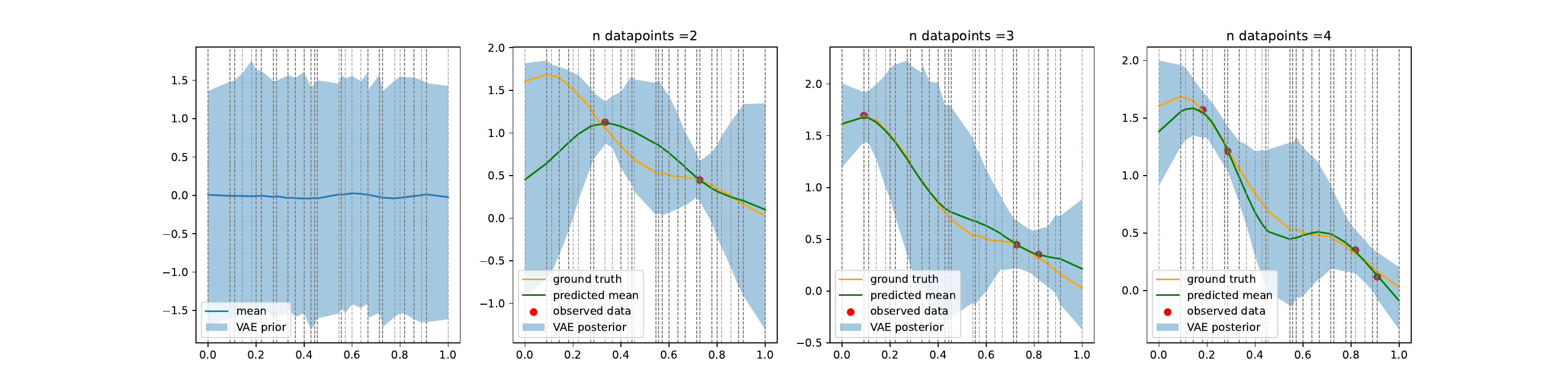}
  \caption{Results of MCMC inference on an irregular grid on noisy GP data by using the trained VAE priors $f_\text{VAE}$. The  posterior mean of our model is shown in green with the $95\%$ credible intervals shown in blue. Quality of the estimation improves with the growing number of data points.}
  \label{fig:irreg_inference}
\end{figure}

Results of our approach on an irregular one-dimentional grid. Figure~\ref{fig:irreg_draws}(a) shows samples of priors from the original GP, evaluated on the irregular grid, Figure~\ref{fig:irreg_draws}(b) shows priors learnt by VAE and Figure~\ref{fig:irreg_inference} shows inference results for different number on observed datapoints.

\section*{Scottish lip cancer dataset}
\begin{figure}[htbp]
   \centering
   \vspace*{-0.5cm} 
  \subfloat[] {
  \includegraphics[width=0.48\textwidth]{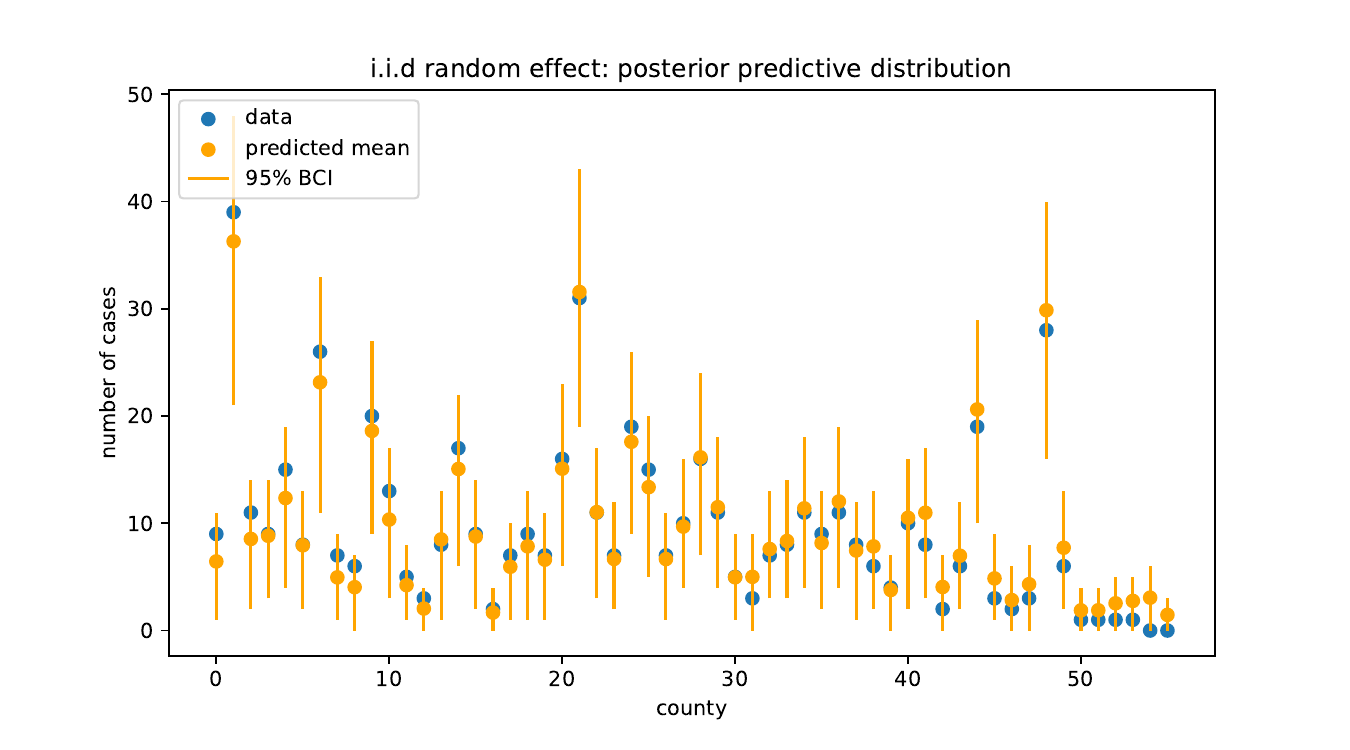} 
} 
\subfloat[] {
 \vspace*{-0.1cm} 
  \includegraphics[width=0.48\textwidth]{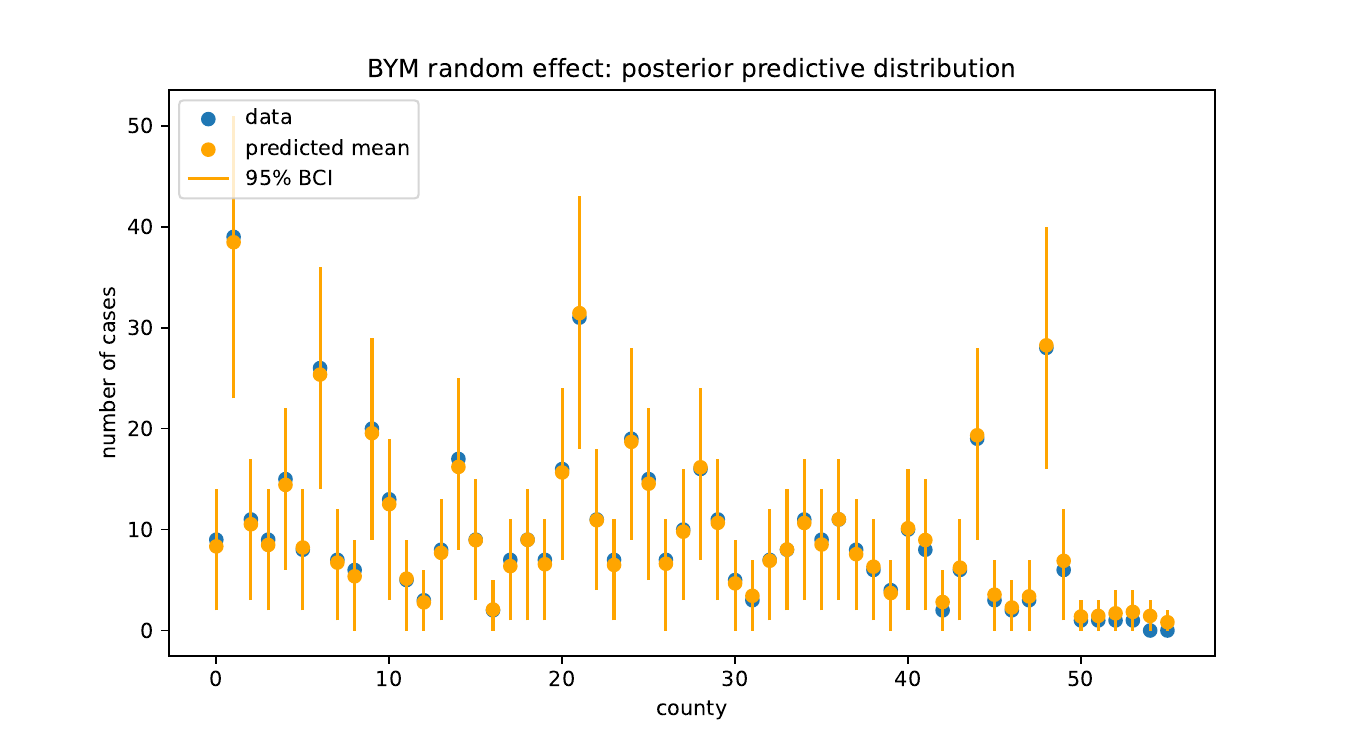}
} \\
\subfloat[] {
  \includegraphics[width=0.48\textwidth]{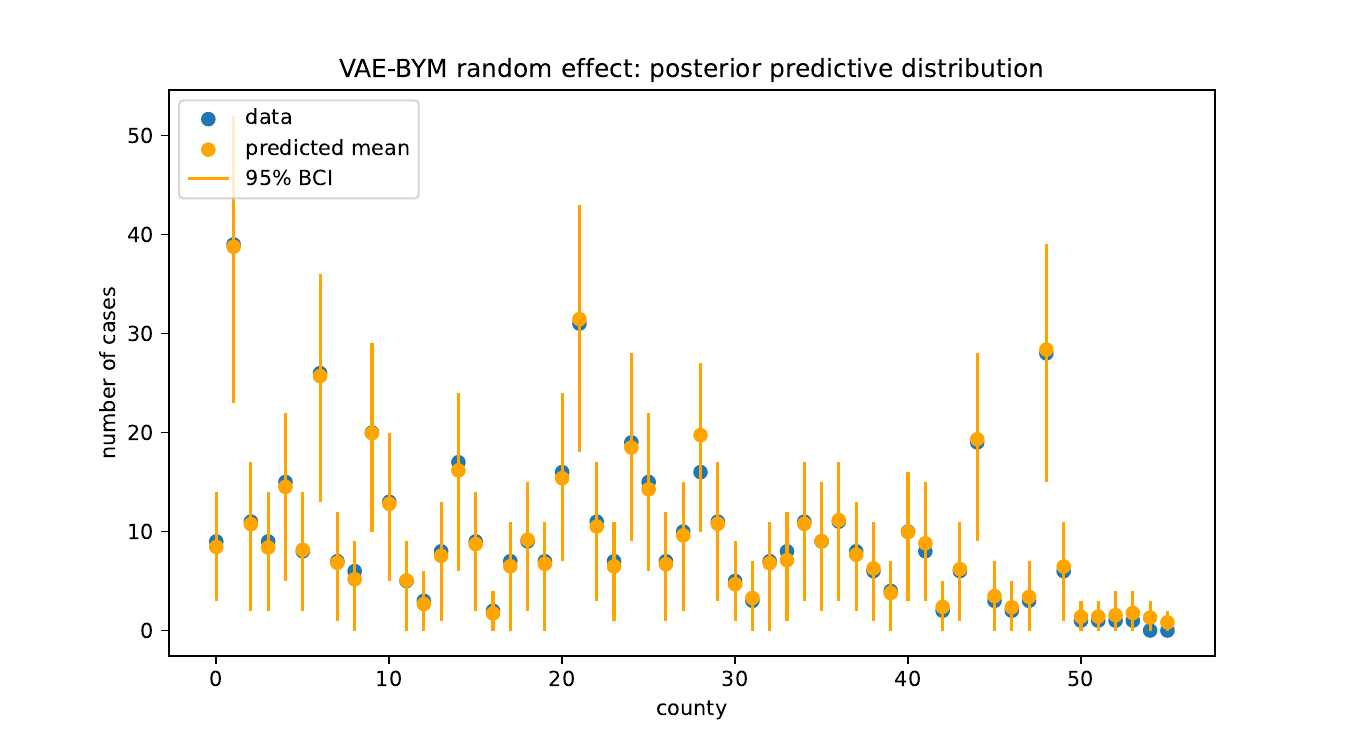}
} 
 \caption{Posterior predictive distributions of the case count data for each out of 56 counties in Scotland. Each posterior distribution is represented by its point estimate (mean) and 95\% Bayesian credible intervals}
 \label{fig:scottish_supplement}
\end{figure}

Here we present results concerned with the Scottish lip cancer dataset, produced by models with i.i.d., BYM and VAE-BYM random effects. Figure ~\ref{fig:scottish_supplement} shows posterior predictive distributions, produced by the three models when all of the available data was used to fit the models. We observe that the i.i.d. model already achieves a relatively good fit. BYM and VAE-BYM models are able to capture the remaming spatial dependence and are very similar between themselves.

\section*{HIV prevalence in Zimbabwe dataset}
\begin{figure}[!htb]
   \begin{minipage}{0.48\textwidth}
     \centering
     \includegraphics[width=.7\linewidth]{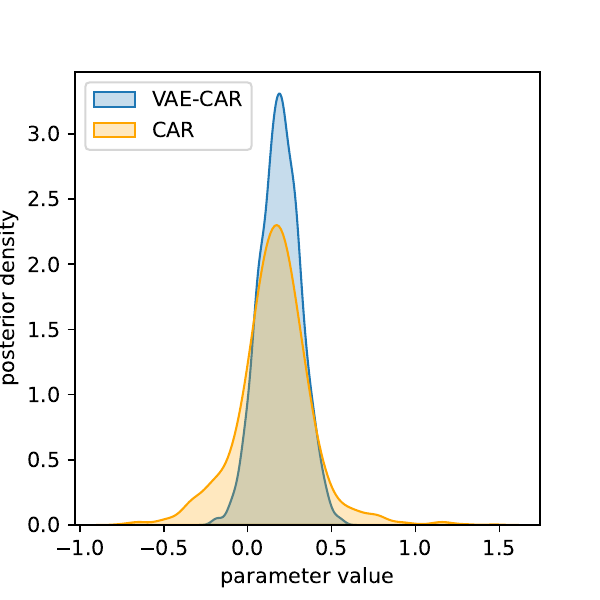}
     \caption{Posterior density of the spatial random effect for one area obtained during MCMC inference on HIV prevalence data from Zimbabwe.}\label{fig:zimb_dens}
   \end{minipage}\hfill
   \begin{minipage}{0.48\textwidth}
     \centering
     \includegraphics[width=.7\linewidth]{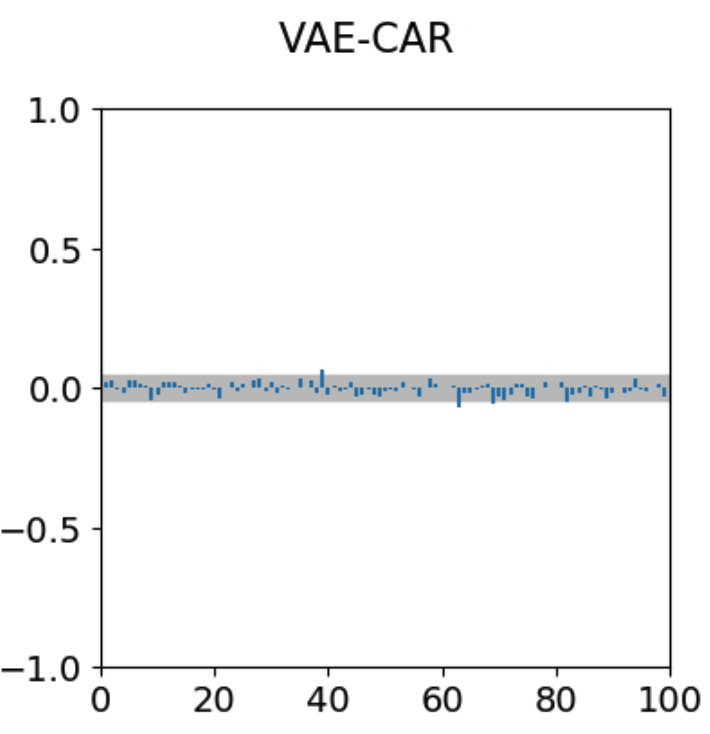}
     \caption{HIV Zimbabwe prevalence data. Autocorrelations of MCMC samples of a spatial random effect modelled via the VAE-CAR model.}\label{fig:zimb_autocorr_vae}
   \end{minipage}
\end{figure}

\begin{figure}[!htb]
\centering
  \includegraphics[height=6.0cm]{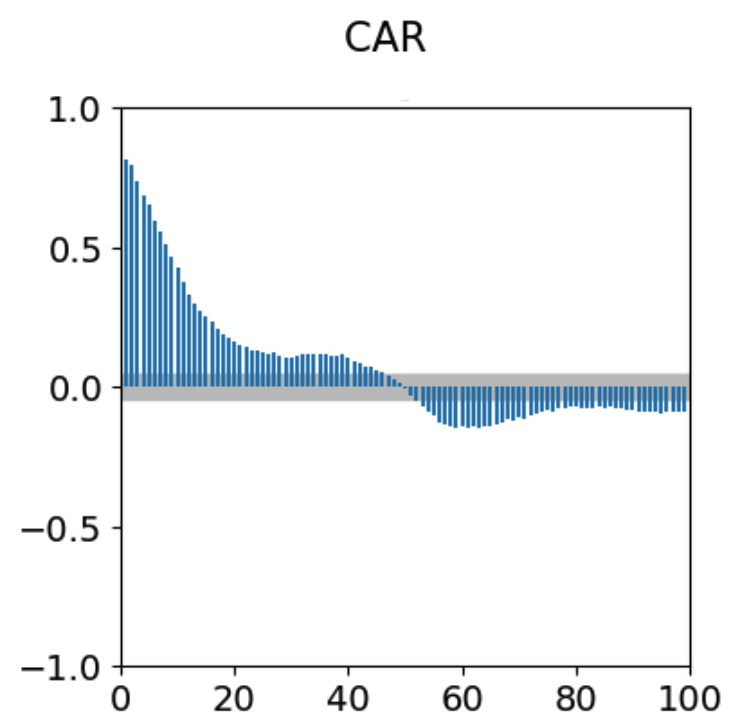}
  \caption{HIV Zimbabwe prevalence data. Autocorrelations of MCMC samples of a spatial random effect modelled via the CAR model.}
  \label{fig:zimb_autocorr_car}
\end{figure}

Here we present results of the study using HIV prevalence data in Zimbabwe: posterior density of the spatial random effect for on earea during MCMC inference (Figure~\ref{fig:zimb_dens}), autocorrelations of MCMC samples of a spatial random effect modelled via the VAE-CAR model (Figure~\ref{fig:zimb_autocorr_vae}) and autocorrelations of MCMC samples of a spatial random effect modelled via the CAR model (Figure~\ref{fig:zimb_autocorr_car}).



\end{document}